\documentclass[11pt]{article}
\usepackage{caption}
\usepackage{subcaption}
\usepackage{amsfonts}
\usepackage{bbm}
\usepackage{amsmath}
\usepackage[ruled,vlined,linesnumbered]{algorithm2e}
\usepackage[section]{placeins}
\usepackage{epsfig}
\usepackage{amssymb}
\usepackage[numbers]{natbib}
\usepackage{graphicx}
\usepackage{amsthm}
\usepackage{booktabs}
\usepackage{etoolbox}
\usepackage{adjustbox}
\usepackage[round-mode=places,detect-weight=true,detect-inline-weight=math]{siunitx}

\theoremstyle{definition}
\newtheorem{definition}{Definition}[section]

\SetKwInput{KwInput}{Input}                
\SetKwInput{KwOutput}{Output}              

\begin{document}

\title{A Computational Theory and Semi-Supervised Algorithm for Clustering}

\author{Nassir Mohammad\thanks{Email: nassir.mohammad@airbus.com} \thanks{Project code is available at https://github.com/M-Nassir/clustering} \\ 
\emph{Cyber Innovation Lab, VCX, Airbus, Newport, UK}
}
\maketitle

\begin{abstract}
A computational theory for clustering and a semi-supervised clustering algorithm is presented. Clustering is defined to be the obtainment of groupings of data such that each group contains no anomalies with respect to a chosen grouping principle and measure; all other examples are considered to be fringe points, isolated anomalies, anomalous clusters or unknown clusters. More precisely, after appropriate modelling under the assumption of uniform random distribution, any example whose expectation of occurrence is $<1$ with respect to a group is considered an anomaly; otherwise it is assigned a membership of that group. Thus, clustering is conceived as the dual of anomaly detection. The representation of data is taken to be the Euclidean distance of a point to a cluster median. This is due to the robustness properties of the median to outliers, its approximate location of centrality and so that decision boundaries are general purpose. The kernel of the clustering method is the perception anomaly detection algorithm, resulting in a parameter-free, fast, and efficient clustering algorithm. Acknowledging that clustering is an interactive and iterative process, the algorithm relies on a small fraction of known relationships between examples. These relationships serve as seeds to define the user's objectives and guide the clustering process. The method then expands the clusters accordingly, leaving the remaining examples for exploration and subsequent iterations.  Results are presented on synthetic and realworld data sets, demonstrating the advantages over the most popular unsupervised and semi-supervised clustering methods.
\end{abstract}


%
\FloatBarrier
\section{Introduction} 
\label{section:introduction} 

Grouping is a natural process carried out by humans in their interaction with the environment and is particularly considered fundamental when interpreting and understanding visual information. This is exemplified in the theories of human vision, and in particular the Gestalt Theory of Psychology where partial gestalts (local groupings) are taken to repeatedly fuse to culminate in global perceptions \cite{Morel07}. It is thus natural to see that grouping---otherwise known as cluster analysis in the field of machine learning---is ubiquitous and appears in practically all disciplines where data is collected. This includes malware clustering in cybersecurity, topic clustering in document processing, fraud analysis in finance and grouping gene expressions in molecular biology. The wide range of subjects highlights the importance of cluster analysis; specifically for the exploration, representation and understanding patterns of interest.

However, clustering data into meaningful groups is a challenging problem. Not least because even defining and framing the problem so as to state precisely what is to be computed is fraught with difficulties. Often the problem is vaguely posed as finding \emph{natural} groupings of the data where within group points are \emph{similar} with respect to some measure like distance, density or probability, yet \emph{dissimilar} from other such formations. However, the problem statement also needs to make explicit that some observations may be noise or anomalies and not belong to any meaningful group, or be altogether a completely unknown group, thus requiring more nuanced decision boundaries and not the forceful lumping of all examples into some groups. 

Another factor that leads to challenges is the type of learning encountered in cluster analysis---machine learning algorithms generally fall under the categories of supervised, unsupervised and semi-supervised. In supervised learning labels of the data are all available and the goal may be to learn a classifier or a predictor. In unsupervised learning data is unlabelled and is a scenario most widely encountered in the realworld because most data is unlabelled and acquiring such labels can be costly (or even unnecessary). The term unsupervised is also often interpreted to mean that the algorithms run without user assistence and interaction; here an algorithm returns what it deems to be the best result, or a number of different results are made available for the user to select according to their goal. The semi-supervised learning category normally assumes that some small fraction of labelled data is available, or only the labels of a particular class; such information can help guide, validate and improve the methods. 

The cluster analysis problem usually falls under the category of unsupervised learning and is used to find patterns of interest; however, deciding what is the ideal way to formulate and guide (optimisation) algorithms is difficult, as well as the measuring of performance. This has resulted in many heuristic based algorithms for which users are left to supply critical parameters that cannot be automatically reliably inferred from the data. Instead, such parameters may be estimated by data exploration, visualisation, sampling and labelling. Thus, it seems reasonable to question whether these algorithms are \emph{truly unsupervised}; instead placing them into the category of semi-supervised learning where even though labelled data is not used at the outset, the user is providing some additional information to the algorithm that is obtained through human intuition and experience of the specific data or domain. Interestingly, such information can often be interpreted as constraints on the possible data partitions or clusters that can be returned by the algorithm, where the parameters specify the resolution at which the process is to be carried out. 

In more recent years clustering has also become explicitly semi-supervised through the ideas of constrained clustering where typically samples of labelled data or relationships are leveraged in some modified version of the $k$-means algorithm \cite{WagstaffCRS01}. This form of clustering is practical and interactive since in many scenarios some data or knowledge of clusters is available and supervision can take the goals of the user into consideration at multiple steps. Furthermore, this specification of constraints for clustering is important because without guidance or additional external information (that is unavailable from the data), the algorithm cannot know the criteria by which to cluster, nor the level of resolution that is commensurate with the goal. 

In the present work both the unsupervised and semi-supervised approaches to clustering are of interest---leaving aside a strict definition. The former usually has benefits of ease of use while the latter can use explicit prior information from the user to guide or constrain the clustering to solutions that are desired amongst the vast number of possibilities.

\subsection{Prior Art}
\label{section:priorArt}
There are several general approaches in unsupervised clustering which include hierarchical, density, spectral, information-theoretic, probability and distance to the centroid based methods. For each approach, together with a representation of the data, many algorithms have been developed that are usually modifications of the fundamental idea so that certain deficiencies are claimed to be overcome. However, this normally introduces complexities that result in an overall less practical algorithm in terms of additional runtime, storage or parameters. It is thus important to note that even with the flux of new algorithms over the past decades, the most widely used and cited algorithms are still agglomerative clustering, DBSCAN, Gaussian mixtures and $k$-means. 

No single algorithm can work best across all data sets and for every data set that an algorithm particularly excels at one can construct another where it fails. However, there are still properties an algorithm may have that are generally problematic. These include: the setting of unintuitive and data specific parameters that can be nigh on impossible to specify in advance and difficult to do so even after data exploration; high runtime and storage requirements with increasing number of observations or dimension of the data; the assignment of anomalies to a cluster when they are clearly alien to the group; the inability to predict on new data points given the current clustering; and being unable to produce good clustering results across as many varieties of data sets and domains actually encountered in the realworld as possible.

For example, in hierarchical clustering such as the agglomerative method \cite{Ward63}, users are required to specify the granularity at which clusters are to be taken. This has the advantage that clustering can be considered at different resolutions or levels but requires nontrivial (and perhaps unintuitive) specification on the users part. There are also a number of linkage methods to select from that can give different results. The algorithms' runtime and storage requirements are often prohibitively expensive in practice, with time complexity scaling as $O(n^2logn)$ and space complexity as $O(n^2)$, as the number of examples increases. This is largely due to the need to store and repeatedly update the full distance matrix after each cluster merge, which adds significant computational overhead. Another important aspect to note is that all merges are final in the building of clusters. This can be problematic when for example, there are anomalies present, since the grouping cannot adjust or correct later to a better solution. Furthermore, the hierarchical method does not natively associate a score with an observation or predict the cluster label of a new observation given the current clustering, this restricts its utility somewhat unless additional classifiers are applied on top.    

\begin{figure}
	\includegraphics[width=\linewidth]{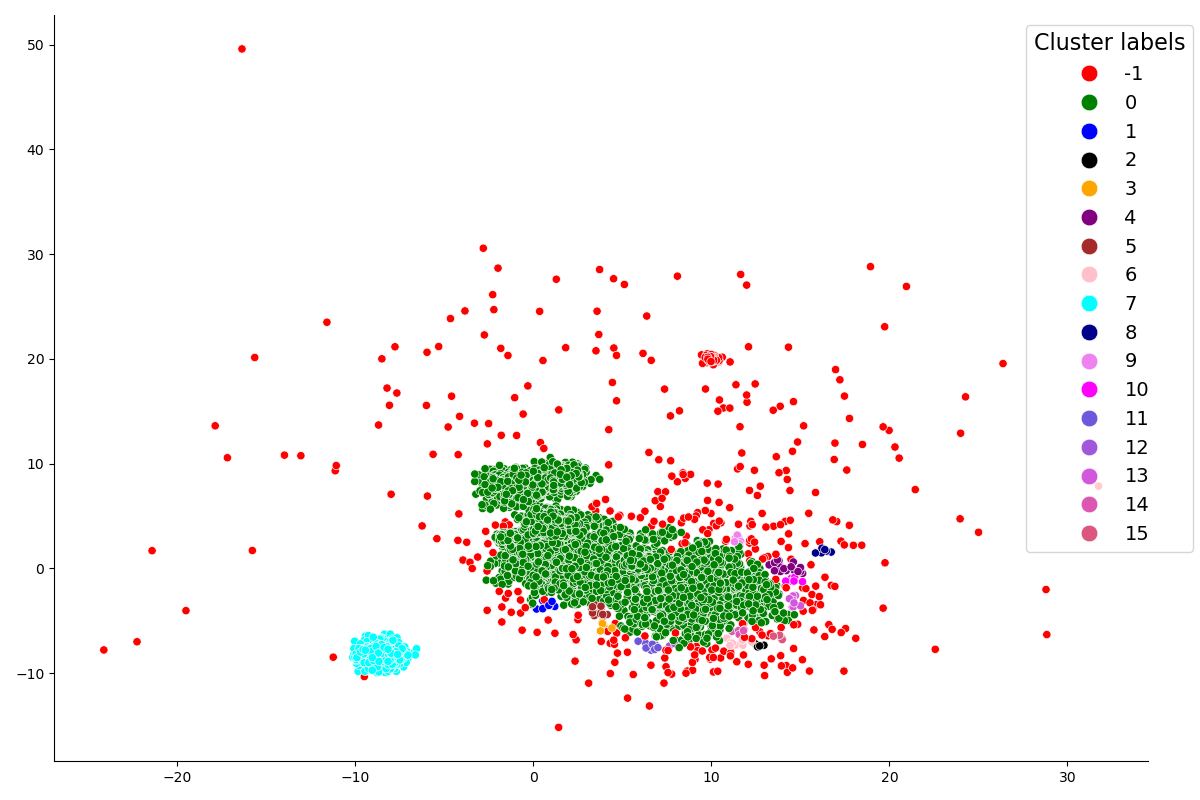}
	\caption{Results of the DBSCAN algorithm with default parameters. Most of the clusters are reported to be a single large cluster while many outlying points are highlighted as anomalies.}
	\label{fig:2d_gaussian_dbscan}
\end{figure}

Density based methods work on the premise that clustering locates regions of high density that are separated from one another by regions of low density. The most popular method is DBSCAN \cite{ester1996density}, which can efficiently cluster arbitrarily shaped data distributions with $O(nlogn)$ runtime and $O(n)$ memory, in typical low-to-moderate dimensional settings, while also potentially detecting anomalies.  However, in high-dimensional data, DBSCAN can become significantly slower—often approaching $O(n^2)$ complexity-due to the increased computational cost of neighborhood queries, as finding $\varepsilon$-neighbors becomes less efficient and more computationally expensive. Moreover, it requires the specification of parameters that are highly sensitive and difficult to tune in practice even when using methods such as the ‘elbow’ method for parameter selection. Consequently, it can fail to give good results in even what appear to be simple data sets if clusters are too close to one another or there are large differences in densities. The method also benefits from post-processing to remove small groups of data that may get incorrectly classed as clusters, and it has been found in private experiments that its anomaly detection results can be unsatisfactory (see Figures \ref{fig:1d_DBSCAN} and \ref{fig:2d_gaussian_dbscan} for examples). Additionally, DBSCAN does not natively provide a cluster association score for every observation nor does it offer predictive ability over new data points; the latter being handled by rerunning the algorithm again with new data. 

Probability-based models aim to fit statistical distributions to the data and estimate their parameters, with Gaussian Mixture Models (GMMs) being among the most widely used. When the underlying assumptions are met, GMMs can yield strong clustering performance, providing not only group assignments but also probabilistic scores indicating the degree of association between data points and clusters. By modelling clusters as distributions, GMMs allow for interpretable, ellipsoidal cluster structures and offer a degree of flexibility across different types of data. However, the method relies on the iterative Expectation-Maximisation (EM) algorithm to estimate the distribution parameters, which can be computationally demanding. Although the cost of each EM iteration is approximately \( O(nkd^2) \)—with \( n \) the number of data points, \( k \) the number of mixture components, and \( d \) the dimensionality—the total number of iterations needed for convergence is highly variable and can grow with data complexity. As a result, the overall runtime is often substantial, particularly for large or high-dimensional datasets. The quadratic scaling with dimension further exacerbates this issue. In addition, GMMs are sensitive to the choice of \( k \), and their performance can degrade when data is poorly separated, colinear, limited in quantity, or deviates from Gaussian assumptions. Noise and anomalies are also not natively handled, as GMMs produce a hard partitioning of the data.

Of all the clustering algorithms mentioned thus far, it is still the simple method of $k$-means \cite{macqueen1967, forgy65} that is the most widely known and stands out as the algorithm of choice for practitioners---despite it being over $50$ years old. It does however have several drawbacks which has resulted in many extensions, but most of which have not gained traction for practical use because of the additional complexities they introduce without necessarily significantly better results. Since it is a centroid based method, and because of its widespread appeal and use, comparisons in the present paper will lean towards $k$-means because any algorithm that can better or complement it will be a significant contribution to the field of cluster analysis. 

$k$-means has a number of important properties that have made it popular. Foremost is its simplicity to understand and to implement, and it is very fast to compute in practice---both with increasing feature dimensions and the number of examples. Its time complexity is $O(tknd)$ ($t$: number of rounds, $k$: number of clusters, $n$: number of observations, $d$: dimension of data) and so it is linear with the number of examples in the data. It's speed also makes it suitable for interactive use which is important for exploratory work. The space complexity is also low, being $O((n+k)d)$. $k$-means being a centroid based method also benefits from being able to natively score points and predict upon new data by simply using the distance to the centroid of the parent cluster; both of which can be beneficial in applications. Like other algorithms it requires parameters, principally the value $k$ specifying how many clusters the algorithm is to return. Indeed, too few will result in some groups being clumped together as one, while too many will result in desired clusters being separated into sub-clusters. However, users can intuitively understand this parameter and in specific problems are able to set it in advance, estimate it using the `elbow' method or gauge it interactively from the clustering results. Thus, although the ideal solution is parameter-free, having only a single intuitive parameter alleviates the burden somewhat and makes $k$-means appear relatively straightforward to use compared to other algorithms.

However, $k$-means is sensitive to the initialisation centroids, resulting in it getting stuck in local optima so that multiple runs with random initialisation is sometimes recommended. However, this may not necessarily work well depending on the data and number of clusters specified. Thus, another approach includes taking samples of points and clustering them using hierarchical clustering where the centroids of those clusters are used as seeds. This approach may only work well if the sample size is small due to the computational complexity of hierarchical clustering and if $k$ is relatively small compared to the sample size. Note that the number of clusters still requires specification and its empirical success remains unclear. Other important and useful extensions include $k$-means$++$ and bisecting $k$-means. The former picks centroids incrementally where new centroids are taken to be a random one that is far from its closest centroid and based on a probability measure that is proportional to the square of its distance. The latter splits the data into two clusters, and repeatedly operates on one of the clusters until $k$ clusters have been produced. Interestingly this method also yields a hierarchical clustering by recording the sequence of clusterings produced as $k$-means bisects each cluster.  

The optimisation criteria and the computation of centroids using local means makes $k$-means sensitive to anomalies which can lead to suboptimal partitions of the data. However, detecting and removing anomalies first is a challenging problem in of itself. In the case of clustering followed by anomaly removal, the sequential combination requires further analysis and practical investigation due to the distorting effect anomalies can have on the initial clustering. Another couple of methods similar to $k$-means, called $k$-medoids and $k$-medians have been proposed to overcome the problem of sensitivity to anomalies. The former takes an actual point to represent a cluster rather than the mean, while the latter uses medians as representatives. However, both are generally computationally more expensive than $k$-means without significant improvement in results. 

Closely related to the problem of anomalies is that $k$-means can give undesirable partitions in the presence of clusters with different sizes or densities. Additionally, because it is a partitioning algorithm it will always place an example into a cluster even if it is an anomaly, a cluster of anomalies or in fact belongs to another unknown cluster. Thus, it is important to differentiate between algorithms that return clusters (such as DBSCAN) and partitioning algorithms since the term clustering is overloaded. It is usually the case that when users want to carry out cluster analysis, they desire groups of data that are similar to be returned while excluding that which does not belong to the group. This has implications with regard to \emph{what} should be computed in the problem of clustering data and whether $k$-means is merely approximating what is desired of an algorithm. The partioning problem is also apparent when the dimensionality of a data set is $>=10$ and the number of desired clusters is $k>=20$; in this situation the $k$-means algorithm often converges with one or more clusters which are either empty or contain very few data points \cite{constrained_kmeans_bradley}.  

A criticism often levied at $k$-means is its inability to cluster certain complex grouped data since it constructs hyperspherical clusters (or hyperellipsoidal clusters when using the Mahalanobis distance). While this is a point of criticism, it may in fact be a contributing factor for its widespread and general use since it does not follow the detailed distribution of the data and potentially suffer from overfitting. Indeed, it is assumed in this paper that for many realworld data sets, desired cluster groupings are roughly hyperspherical or are separate enough that such a clustering assumption works well in practice. 

$k$-means has also been the basis of many semi-supervised clustering algorithms, with perhaps the most popular being constrained $k$-means \cite{constrained_kmeans_bradley}, seeded $k$-means \cite{BasuBM02}, $COPK$-means \cite{WagstaffCRS01}, $PCK$-means \cite{basu_pckmeans_2004} and MPCKMeans \cite{BilenkoBM04}. These algorithms work in a number of different ways. Constrained $k$-means explicitly adds $k$ contraints to the underlying clustering optimisation problem requiring that each cluster have at least a minimum number of points in it. Seeded $k$-means initialises the centroids using pre-defined seeds, which can improve the clustering results when domain knowledge is available. A variant of this we call constrained seeded $k$-means fixes the supplied seeds and runs clustering on the non-seed data. $COPK$-means operates by assimilating hard constraints from the user (e.g., must-link and cannot-link) that cannot be violated while carrying out $k$-means clustering. Thus, the constraint labelling is assumed noiseless. $PCK$-means also incorporates constraints from the user, but these are soft constraints where the algorithm suffers a penalty for violation. Finally, MPCKMeans enhances the constrained clustering approach by combining soft pairwise constraints with metric learning so that individual metrics (Mahalanobis distance) for each cluster are learned which permits clusters of different shapes. 

Other semi-supervised clustering algorithms of interest include semi-supervised DBSCAN, which uses labelled data to help obtain suitable parameters that characterise $each$ of the different clusters \cite{SSDBSCANLelisS09}. Semi-supervised Gaussian Mixture Models (GMMs) adapt the probabilistic clustering framework by integrating label information—either in the form of partially labelled data or pairwise constraints \cite{ShentalBHW03}. Semi-supervised spectral clustering enhances traditional spectral clustering by integrating prior knowledge, such as pairwise constraints or labelled data, into the similarity graph or its Laplacian matrix. This is commonly done by modifying the affinity matrix or adapting the graph partitioning objective to satisfy the imposed constraints \cite{YuS01, KamvarKM03}.

All the semi-supervised algorithms discussed thus far have claimed to improve clustering performance over the underlying unsupervised base versions. However, although the application domain is important (since oftentimes some labelled data or constraints are available), these methods have received less attention in practice. This is likely due to several factors: many of the original studies were conducted on relatively small datasets with a lack of extensive benchmarking against strong baselines; the algorithms themselves tend to be computationally expensive, especially on medium-to-large datasets; and available implementations are often research-grade, with few mature, well-supported libraries for practical use. Moreover, the performance gains are frequently modest unless an impractically large number of constraints is provided. Finally, any practical limitations inherent in the base clustering algorithm typically persist in its semi-supervised variants, further limiting their appeal in real-world applications.

In recent years, deep learning-based clustering methods have garnered significant interest for their ability to handle complex, high-dimensional, and unstructured data. Unlike traditional clustering techniques, which typically operate on the original feature space or rely on manually engineered features, deep learning models can learn more meaningful data representations prior to clustering \cite{TianGCCL14, PengXFYY16}. Notably, substantial performance gains are often achieved when feature extraction and cluster assignment are jointly optimised within an end-to-end framework. A seminal example of this approach is Deep Embedded Clustering (DEC) \cite{XieGF16_DEC}, which first uses a deep autoencoder to learn a low-dimensional latent representation of the data. The decoder is then discarded, and the encoder is fine-tuned by simultaneously refining the latent features and optimising cluster assignments. This is done by minimising a clustering-oriented loss, typically based on the Kullback-Leibler (KL) divergence between a soft cluster assignment distribution and a target distribution that encourages more confident cluster assignments. Another  method, Deep Clustering Network (DCN) \cite{Yang2017_DCN}, jointly optimises a deep autoencoder and a k-means clustering objective within the latent space, balancing reconstruction accuracy and clustering performance. More recently, SHADE (Structure-preserving High-dimensional Analysis with Density-based Exploration) \cite{SHADE_2024} has extended deep clustering by introducing a density-connectivity loss that enables the discovery of arbitrarily shaped, non-Gaussian clusters. It automatically detects both clusters and noise without needing predefined cluster counts, making it appealing in complex, unsupervised scenarios. While deep clustering methods have the potential to outperform traditional clustering approaches, this advantage must be validated through comprehensive experimental evaluation, since they often still utilise unsupervised learning algorithms. Moreover, any performance gains are often offset by increased model complexity, higher computational cost, and a lack of robust, production-ready implementations.

Semi-supervised adaptations of deep learning-based clustering methods have also been proposed, typically incorporating constraints into the objective function to leverage prior knowledge and guide the clustering process \cite{GonzalezAlmagroPPCG25}. For instance, Semi-Supervised Deep Embedded Clustering (SSDEC) extends DEC by integrating pairwise constraints \cite{RenHDPHX19}. Another method, Deep Constrained Clustering (DCC) \cite{ZhangBD19}, provides a flexible framework that can incorporate various types of constraints — including pairwise relations, instance-level difficulty, and cluster size balance — by employing differentiable surrogate loss functions. In contrast, SHADE-CC \cite{GeneticGonzalezAlmagro21} is not based on deep learning; instead, it leverages evolutionary optimisation, specifically the Success-History based Adaptive Differential Evolution algorithm, to identify high-quality clusterings under constraints. While evolutionary methods like SHADE-CC are generally not designed for large-scale data, they can be effective on small to medium-sized datasets with complex constraint structures.  

Finally, it is worth noting that several works have attempted to combine clustering with outlier detection, although these tasks are often treated independently. This separation can be problematic, as outliers can substantially distort cluster structures, while reliably identifying outliers remains a challenging problem in itself. To address this, some methods seek to integrate outlier detection directly within the clustering process. A well-known example is DBSCAN, which naturally identifies low-density points as noise or outliers while forming clusters. However, in DBSCAN, outlier detection is a secondary outcome rather than a primary objective, and its effectiveness can be inconsistent depending on parameter settings and data characteristics. An alternative approach is k-means--(minus minus) \cite{ChawlaGionis2013}, an extension of the standard k-means algorithm that iteratively excludes a fixed number of points considered outliers based on their distance to the nearest cluster centroid. While this improves robustness, it introduces an additional parameter—the number of outliers to remove per iteration—which (as stated by the authors) is very difficult to determine in practice. Another notable method, Clustering with Outlier Removal (COR) \cite{LiuCOR2018}, jointly optimises clustering and outlier detection by performing multiple runs of k-means--. These runs produce varied partitions, which are then used to map data points into a binary space, improving resilience to outliers. The final clustering is obtained by optimising a unified objective based on Holoentropy. Although COR has claimed good performance, it also requires careful parameter tuning and can be computationally expensive on large datasets due to the repeated k-means-- runs, binary transformations, and optimisation procedure.

\subsection{Contributions}
\label{section:contributions}

Data clustering is intended to be carried out by a computer due to the size, number, diversity and speed at which information needs to be processed. However, oftentimes there is a comparison with the human ability to cluster sensory stimuli or data (particularly in low dimensions) which is carried out innately and effortlessly. Thus, to reach an ideal solution it seems pertinent to understand and learn from how humans might address or even solve the problem---particularly the diversity aspect and speed at which it is conducted.  

Our starting point will be to consider clustering systematically as an information processing task and addressing it using David Marr's \cite{Mar82} three levels of analysis---particularly the computational theory and algorithmic levels. This approach disciplines us to ask the correct questions at the appropriate levels of explanation so that we can better understand the overall task. Thus, beginning with the top level of computation we can ask in general and precise terms what is to be computed---the inputs and outputs---and why it is appropriate, and what realworld constraints should be incorporated to make clustering tractable. Then we can consider the next level of processing to derive representations of the data examples, the inputs and outputs, the nature of the ill-posed problem, and how we might bring in additional assumptions to develop an algorithm for carrying out the transformation from input to the desired output. 

The present work builds a clustering algorithm from the ground up by following the above analyses and using human visual perception as inspiration. This approach was successfully used by \citet{nassir2021anomaly} in his development of the  anomaly detection algorithm called \emph{perception}, which will form a core component of the clustering method. Indeed, clustering is considered the complement of anomaly detection (sensing distinctness), where clusters are deemed to be sets of data points that do not contain any anomalies with respect to a chosen grouping principle and measure. Mohammad's definition of an anomaly is used, and data points are represented as the distance to the median of the parent group so that hyperspherical clusters are formed in order to have a solution that is generally applicable and practically useful like the $k$-means algorithm  (while accepting failure in other and non-hyperspherical distributions). Furthermore, anomaly detection being the core of the approach naturally results in fringe points, isolated points or small groups being labelled as anomalies so that clustering---and not a partitioning of the data---is carried out.

Note that clustering is done to discover meaningful groupings of the data or interesting patterns that may not have been previously known. However, without embedding or conveying to an algorithm what is desired or considered meaningful, an algorithm cannot know which grouping of the data is appropriate. Indeed, any random clustering could be returned and justifiably so since constraints or requirements have not been provided. This resonates with the idea from human vision studies that intelligence identifies the appropriate resolution for the present utility, with perception iconicising the visual input at that appropriate level. Most algorithms implicitly assume some desired properties of the clusters in addition to requiring parameters, these naturally constrain the possible solutions. In the case of $k$-means it is that clusters are hyperspherical in nature and of similar size and density, and the data is without noise or anomalies, with a specified number of clusters; the computational problem being to minimise the within partition sum of squared error. While for DBSCAN regions of high density are considered clusters where the output is constrained by the two parameters specifying a radius to compute densities and the minimum number of points to be in that region. In the semi-supervised algorithms such as $COPK$-means, the base $k$-means assumptions are held, and in addition the labelled examples or pairwise constraints are provided to help the clustering approach the users desired outcome.    

The clustering method presented in this paper is parameter-free but assumes an interactive process between human and machine, where users must provide a small amount of labelled data that approaches a constant with increasing size of the data per cluster, so that seed members of \emph{known} groups are obtained. This little information is enough to constrain and guide the algorithm to grow meaningful clusters and carry out both local and global anomaly detection. As users interact with the clustering outputs, further observations can be sampled, analysed and discovered, while existing clusters may be modified. The method---being based on the perception algorithm---is fast, intuitive and light on memory usage. Furthermore, because there is no universally best clustering algorithm, a framework is sketched for dealing with other useful representations of data in order to tackle different data distributions. 

To detail the contributions of this work, the rest of the paper is laid out as follows: Section~\ref{section:computational_theory} introduces the computational theory underpinning the approach and presents the proposed parameter-free, semi-supervised clustering algorithm. Section~\ref{section:example_results} illustrates the algorithm’s performance on synthetic one- and two-dimensional datasets to provide an intuitive understanding of its behaviour. Additional results on publicly available data sets are also provided to demonstrate realworld applications. (Readers are encouraged to visit the GitHub repository linked in the footnote on the first page of the present paper to explore additional examples and experiment with the method.) Finally, Section~\ref{section:conclusion} concludes the paper and outlines directions for future work.

\section{A Computational Theory for Clustering}
\label{section:computational_theory}

The goal of the present work is to draw insights from the human visual process of object grouping and develop an algorithm capable of automating clustering tasks on large datasets. In the case of complete automation we will have an unsupervised solution, whereas when we permit a small amount of human involvement (labelling data) or interaction (iterative evaluation) we will have a semi-supervised solution. 

Human vision is considered to be an example of complex information processing for which \citet*{Mar82} introduced the idea of a three level analysis. His core theme was that vision is not governed by a single explanation, equation or level of understanding, but that each visual task requires its own multiple levels of analysis which form a top-down hierarchy. This approach has significantly impacted the field of computational neuroscience because it helps delineate what is to be computed and whether it satisfies assumptions obtained from the realworld and with what is desired of the task, from how one might carry out the computation and its implementation. This section will describe the levels of analysis as applied to the problem of clustering, and present an algorithm for efficient computation.

\subsection{Tri-level Hypothesis}
\label{subsection:tri-level}

The top level in Marr's analysis is called the \emph{computational} level and deals with \emph{what} the computation is, or the goal of the task, and \emph{why} it is appropriate; this is interpreted at both the general and specific levels. At the general level we answer what information is extracted and why it is useful, while at the specific level we state exactly what is to be computed from input to output and what realworld constraints or assumptions are generally true of the world and powerful enough, to allow a process to be defined and the desired outcome to be achieved. The constraints help decide whether the actual function is the one that should be computed, and why this computation is specified instead of some other for the task at hand. Finding such constraints that impose themselves naturally to sufficiently allow a unique solution is a true discovery of permanent value even if it is not internally verifiable. 

The second level in the analysis is the \emph{algorithmic} level and is concerned with \emph{how} the computations associated with the first level are to be carried out. This is broken into two parts. First is the specification and representation of the data input and output, and second is the actual algorithm that carries out the transformation. Note that many representations can be put forward, and each can make some aspects explicit and relegate others to the background---this has significant bearing on the choice of algorithm. Indeed, for each representation many algorithms can be chosen and each algorithm may have certain desirable and undesirable characteristics according to the specific problem. 

The third level of analysis, known as the \emph{implementation} level, focuses on the physical realisation of the computations. This can involve various systems such as the human brain or specialized mechanical machines; but for our purposes it will be a single computer. Nevertheless, it may be important to consider the potential extensions to distributed implementations in scenarios such as big data applications and distributed devices.  

\subsection{Computational Theory}
\label{subsection:clusteringTheory}
The clustering problem is generally understood to be the grouping of portions of data that are similar in a particular way, but separate from other such groups. The process has important applications in data mining where the output can be used---amongst other purposes---for segmentation, summarisation and classification. The input to the problem is assumed to be in a dataframe format, where examples are represented as rows and numerical feature values as columns. The desired output is a one-dimensional array of labels that identify which cluster each example belongs to and in addition there may be a membership score. Note that a specific cluster label is required for the unknown or anomalous group. Unfortunately, there is not enough information available in the input to get a unique desired output directly and hence \emph{any} partitioning of the data could be returned. The problem is ill-posed, and we require additional external information to constrain the possible solutions; we derive this from the human experience of grouping data points in low dimensisonal spaces. Thus, a key research insight in clustering lies in identifying principled constraints and assumptions that meaningfully reflect the structure of the data and the objectives of the task.

The process of clustering is distinct from partitioning; subsets of data points may be grouped into a cluster but not necessarily all points should be assigned to a cluster. This then provides our first natural and most important constraint that \emph{similar points should be grouped together, but anomalies with respect to every grouping and measure, must not belong to any cluster}. The second assumption is that an \emph{object that is part of a cluster has both a membership and degree of membership}. In other words even within a single group all observations have a level of belonging, e.g., points that are further away from the centroid or occupy less dense regions of space than others.

The removal of what is anomalous---with respect to some characteristic---from a set of data points, leaves us with a group of points that are now deemed similar; hence, the following definition of a cluster is made:	

\begin{definition}[Cluster]
	A cluster is a subset of data points that does not contain any anomalies with respect to a chosen grouping principle and measure.
\end{definition}
This definition needs to be made precise and requires some additional explanation and support in order for computation. The Gestalt groupings play an important role since clustering is often described as finding natural groupings in the data. (Figure \ref{fig:1d_gaussian_y_true} illustrates some examples of `natural' clusters by proximity.) Thus, it can be argued that satisfactory clustering results are not completely aribtrary, in that though a staggeringly large number of possible cluster outputs can be returned, it is only a relative few that are acceptable. Thus, the groupings that we care about and consider natural are limited and are conjectured to include grouping by the gestalt laws such as proximity, good continuity and similarity. These groupings should agree with what the common man sees and believes to be true. \emph{Clustering is in the eye of the beholder} is an old adage, but the eye being human is constrained by only perceiving certain patterns that correspond to prior grouping laws of interest. Assuming that the realworld physically imposes natural groupings of interest, the term anomaly is taken to be that defined by \Citet{nassir2021anomaly}.

\begin{definition}[Anomaly]
A grouping of interest represented by a gestalt law is perceived by the Helmholtz principle when it is unexpected to happen (i.e. its expectation of occurrence is $<1$) in uniform random noise. Any observation that is unexpected to occur with respect to this grouping is perceived, by the same principle, to be an anomaly.
\end{definition}

Anomalies, according to this definition can be computed precisely, and an algorithm under a particular representation of Euclidean distance from the median is provided by \citet{nassir2021anomaly}. This not only provides us with a classification but also a score defining the degree of anomalousness. Hence, clusters being simply the complement of the set of anomalies in the data are also computable and each example has an associated score. This also provides us another constraint: \emph{what is computed must have parameters or thresholds that adapt to the data, and are not fixed unless they are universally useable}. This is reasonable to assume since data sets, and indeed clusters within data sets, can vary in terms of numbers, size, density or shape and an algorithm needs to be able to best handle and adapt to such differences. However, the constraint is unlikely to be universally satisfied across all data sets due to the range of possible data distributions and the specificity of each adaptation.

The clustering process is approached from the perspective of removing anomalies as opposed to forming groups, as it provides a more intuitive and appreciable framework. This viewpoint allows for simpler understanding and handling of the subject and algorithms, emphasising that points classified as not belonging to any group are considered anomalies. 

\subsection{Representation}
\label{section:clustering_representation}
There are many ways in which humans group things, and how things can be represented. For example, there is grouping by shape, size, continuity, distance between objects, colour and any combination of these. Thus, the clustering problem also requires users to specify the goal by which data points are desired to be grouped by, and it is only the grouping that has some utility that is desired. It is assumed that clustering is an interactive process in that a user does not simply run an algorithm and take the output at face value. Rather, checks are carried out, samples of clusters are taken, and the results are intuitively checked for consistency and the data explored further. Hence, the information by which the grouping is desired is obtained by the assumption that \emph{a user is able to provide cluster labels to a tiny fraction of the data in order to seed the required goals of the clustering}. (This is perhaps akin to how a child may view some unknown objects and is then provided only a handful of labelled examples from which it is able to extrapolate and learn the abstract class.) Note that together with the exclusion of anomalies from a group, and the labelling of some data points that belong to a group from which clusters can be grown, another constraint is realised: \emph{unknown clusters of data that have no samples of labels should not be incorrectly assigned a cluster label of another group}. Unknown groups here should simply be highlighted as anomalies with respect to all other groupings, and require uncovering through the users iterative and interactive exploration and labelling of the data. 

Although it may appear that requesting users to supply similar examples is difficult, any clustering algorithm output necessarily requires that the user evaluate the results by looking at samples from the cluster to ascertain whether meaningful groupings have been obtained. This necessarily requires that the user has some domain expertise or knowledge of the features. This part of the evaluation is brought forward so that users are asked in advance to provide a small number of examples that likely belong to the same cluster. The labelling does not have to be comprehensive or completely accurate, since toleration of a minority of mislabelled examples is handled by the ejection of anomalies. 

For a given representation of the data points, certain groupings of interest are highlighted, and others are hidden. The representations often used in cluster analysis include distance from the centroid, density within a radius of a point or distance between two examples. (Note even the notion of distance between two points can be broken down into further subgroups like Euclidean, Manhattan or Mahalanobis). Thus, another constraint that is selected by the user is the choice of representation and measure. For the purposes of clustering in this paper, it is assumed that \emph{the Euclidean distance to the centroid (median) of a cluster is an appropriate measure} by which data points are to be grouped, or equivalently anomalies removed from the set to leave a grouping. This is because it is assumed data is often transformed such that the Euclidean distance works well and is one of the reasons why $k$-means has wide and general applicability. Furthermore, there is no evidence that another metric is better in general. The median is the chosen centroid instead of the mean because of its robustness to extreme deviations and better measure of central tendency. 

Other measures related to grouping laws such as density, good continuity, repetition, etc. are also of importance and may be preferred in certain applications. In such cases a grouping of the data must be chosen. For example, a plane of best fit may be selected if the grouping of interest is best described by good continuity, hence any examples that deviate by an unexpected amount would be deemed anomalous with respect to the group.     

\subsection{Algorithm}
\label{section:clustering_algorithm}
Given a representation of the data points---which is defined as the Euclidean distance from the associated median---and a small subsample of labelled data, this section presents an algorithm for transforming the input into the desired output. The main computational task is to detect and remove anomalies from a set of data points, while simultaneously expanding clusters by incorporating non-anomalous points. Thus, anomaly detection lies at the core of the philosophy and method employed.

Here, we also acknowledge an additional constraint arising from the human ability to quickly perceive clusters in low-dimensional data, as well as the requirement to efficiently compute clusters over large realworld datasets. Namely, that the developed \emph{algorithm must be fast to compute and scale efficiently with increasing numbers of examples and dimensions of the data}. Many clustering algorithms that yield satisfactory results with small datasets or low-dimensional data often struggle to meet these requirements, resulting in unacceptably long computation times. Our approach prioritises computational efficiency to overcome these limitations.  

The clustering method presented in this paper employs the perception anomaly detection algorithm due to several advantageous properties: it is parameter-free, adapts naturally to numerical data, is centroid-based for general applicability, and demonstrates computational efficiency on large datasets. Furthermore, it achieves state-of-the-art performance over metrics such as F1 score, AUC, precision, and recall on real-world data \cite{nassir2021anomaly}. An explanation of the algorithm with a simple sketch is given as follows, while full details can be found in \cite{nassir2021anomaly}. Consider a list of integers $V=[v_1, v_2, \dots, v_W]$ in which to detect anomalies, and assume they have been derived from a binary indicator stream $D$ composed of $1's$ and $0's$ where the stream was divided into $W$ fixed-length windows $w_1, w_2, \dots, w_W$, each of length $L$ (which can simply be assumed to be the largest integer in $V$). We can sum all the indicators in each window to have $V=[v_1, v_2, \dots, v_W]=[sum(w_1), sum(w_2), \dots, sum(w_W)]$, and let $S=sum(V)$ be the sum of \emph{all} the indicators across all the windows.

The a-contrario assumption and model is that all the indicators ($1's$ of total sum $S$) are uniformly and independently randomly distributed across the $W$ windows. Under this model, the quality of interest is the unexpectedly large number of indicators that appear in a window. This is formalised by computing the expected number of $n$-tuples $\mathbb{E}(C_n)$ under the a-contrario distribution as a measure of how anomalous the number of elements in a particular window are. Thus, we are interested in events where the expected value is $<1$, which acts as a natural threshold for whether we expect the event to occur given the data parameters. This threshold is both principled and data-driven: if an event is unexpected under the assumed random distribution but appears in the data, it is, by definition, anomalous. Thus, the general formula for measuring the unexpectedness of each $n$-tuple and flagging an anomaly in $V$ is,

\begin{equation}
\label{equation:anomaly_detector}
\mathbb{E}(C_n) = {S \choose n} \frac{1}{W^{n-1}} <1
\end{equation}  
This calculation gives an immediate binary output on the training data to class each window (value from V) as anomalous or benign. There are no parameters left to the user to specify. The use of expectations handles the probability thresholding problem to immediately arrive at the anomalous events, which can subsequently be ranked with smaller expectations corresponding to more meaningful (unexpected) events. The anomaly detection method easily extends to the multi-dimensional case, and the general training and test process together with actual implementation details is described in algorithms \ref{algorithm:PerceptionTrain} and \ref{algorithm:PerceptionTest}.

\begin{algorithm}
	\SetAlgoLined
	\caption{PerceptionTrain}\label{algorithm:PerceptionTrain}
	\DontPrintSemicolon
	  \KwInput{	  	
		  $X$  - input $l \times m$ dimensional array of numbers \newline
		  $acc$ - decimal place accuracy of input data (default 4) \newline
		  $metric$ - distance metric to be used (default Euclidean) \newline
	}
	\BlankLine
	  \KwOutput{  
		  $S$ - sum of indicators \newline
			$W$ - number of windows \newline
			$multiDimMed$ - median values of each feature of X \newline
			$med$ - median of one-dimensional data \newline
			$\mu$ - mean of each feature of X  \newline
			$\sigma$ - standard deviation of feature of X
	}
	  
	\BlankLine
	\If{$dimension(X) > 1$} {
		$X_s, \mu, \sigma \leftarrow standardise(X)$\; 
		$multiDimMed \leftarrow median(X_s, axis=column)$ \; 
		$X_d \leftarrow distance(X_s, multiDimMed, metric)$\;        	
		$X \gets X_d$
	}
	\BlankLine
	$X_g \leftarrow roundAndMakeInteger(X, acc)$	\;		
	$med \leftarrow round(median(X_g))$\;
	$X_f \leftarrow |X_g - med|$\;
	$S \leftarrow sum(X_f)$\;					
	 $W \leftarrow length(X_f)$\;
	 
	\end{algorithm}
	
	\begin{algorithm}
	\SetAlgoLined
	\caption{PerceptionTest}
	\label{algorithm:PerceptionTest}
	\DontPrintSemicolon
	
	  \KwInput{	
		  $X$ - input $l \times m$ dimensional array of numbers to test \newline
		  $S, W, multiDimMed, med, acc, metric, \mu, \sigma$ - from training stage
	}
	\BlankLine
	  \KwOutput{  
		  $Z$ - array of prediction scores for each example row in $X$ \newline
			$Y’$ - binary array of predictions for $X$ (1: anomaly, 0: normal) 
	}
	\BlankLine
	
	\If {$dimension(X) > 1$} {
		$X_s \leftarrow standardise_{\mu, \sigma}(X)$\;  
				$X_d \leftarrow distance(X_s, multiDimMed, metric)$\;        	
		$X \gets X_d$
	}
	\BlankLine
		$X_g \gets roundAndMakeInteger(X, acc)$\;			
		$X_f \gets |X_g - med|$
	\BlankLine
		Initialise $Y'$ to zero array of length $l$\;
		  \For{$i = 0$ to $l-1$}{
			$n \gets X_f[i]$\;
			$Z[i] \leftarrow -1/S \times (log{S \choose n} - (n - 1) \times log(W))$\;	
			\If {$Z[i] > 0$}{
				$Y’[i] \leftarrow 1$ 		
			}
		}
	\end{algorithm}

The proposed clustering method begins with a small, user-provided subsample of data points with known and mostly accurate cluster assignments which may be drawn randomly from the dataset. These assignments act as seeds to initialise and guide the clustering process, while all remaining points are initially considered anomalies and assigned a label of $-1$. The clusters are then ordered by their compactness, as measured by the minimum sum of squared errors. For each cluster, the perception algorithm is applied to identify potential anomalies within it. Any detected anomalies are ejected from the cluster and reassigned the anomaly label $-1$. The perception algorithm is then re-fit to the updated cluster, and predictions are made over all current anomalies to determine whether any should be added to the cluster. This process over all clusters iterates until the cluster assignments stabilise (which typically occurs after only a few iterations) or a predefined iteration limit is reached. As a final step, the perception algorithm is re-applied to each cluster to compute a membership score for each assigned data point, quantifying its degree of belonging. The full procedure is detailed in Algorithm~\ref{algorithm:Clustering}. (Note that application of the clustering method requires users to standardise the data to ensure features are on a similar scale. Any new data to test the method upon also requires the same standardisation parameters as those used during training.)

The clustering method iteratively assigns points to clusters by fitting and applying the perception anomaly detection method within each cluster. Since the perception has a computational complexity of \(O(n \cdot d)\) for \(n\) points and \(d\) dimensions, running the perception method once per cluster per iteration results in a total cost of \(O(n \cdot d)\) per iteration for the entire dataset. Over a maximum of \(t\) iterations, which is typically small, this leads to an overall complexity of \(O(t \cdot n \cdot d)\). Additionally, the final fitting and scoring step involves running perception on all clusters across the entire dataset, incurring an extra cost of \(O(k \cdot n \cdot d)\), where \(k\) is the number of clusters. Since usually \(k \ll n\), the method scales approximately linearly with respect to the number of points and features, making it computationally efficient and suitable for moderate to large datasets.

The proposed clustering method offers a range of properties that meet our design goals while delivering practical utility. It integrates clustering with anomaly detection, allowing points that do not correspond to any known group to remain unassigned. These unlabelled points can be retained for exploratory analysis, potentially uncovering new or emerging clusters. By employing the Perception anomaly detector, the method adapts naturally to the structure of the data and requires no user-defined parameters, making it simple and robust in practice. It operates with minimal supervision, needing only a small subset of labelled examples to initiate the clustering process. For each assigned point, the method provides both a discrete cluster label and a membership score, indicating the strength of association and supporting nuanced downstream analysis. The approach also generalises well to unseen data: new points can be classified by evaluating their anomaly scores with respect to each learned cluster, enabling consistent and reliable prediction.

\begin{algorithm}
	\SetAlgoLined
	\caption{Semi-supervised clustering}\label{algorithm:Clustering}
	\SetKwInOut{Input}{Input}
	\SetKwInOut{Output}{Output}
	\SetKwInOut{Initialisation}{Initialisation}
	\KwInput{$X$ - Data points to cluster,  \newline
		   $X_l$ - subsample of labelled data from $X$,  \newline
	      $max\_n\_iterations=1000$
		  }
	\KwOutput{
		$L$ - Cluster or anomaly label for each example in X \newline 
		$M$ - Cluster Membership score for each example in X \newline
		}
	\BlankLine
	\Initialisation
	\BlankLine
	Initialise all unlabelled points ($X - X_l$) as anomalies with label $-1$\;
	Initialise all labelled points ($X_l$) to a cluster number\;
	Order the labelled cluster groups by minimum sum of squared error\;
	$counter = 0$\;
	\BlankLine

	\While{cluster changes occur \textbf{and} $counter < \text{max\_n\_iterations}$}{
		\BlankLine
		\ForEach{cluster}{
			\BlankLine
			\tcp{Fit anomaly detector and check for anomalies:}
			$clf = \text{Perception()}$ \;
			$clf.fit(\text{cluster})$\;
			$clf.predict(\text{cluster})$\;
			\BlankLine
			\If{anomalies are found}{
				Remove anomalies from the cluster and assign them to the anomaly category\;
			}
			\BlankLine
			\tcp{Re-fit classifier on cluster without anomalies:}
			$clf.fit(\text{cluster})$\;
			\BlankLine
			\tcp{Examine all anomalous points:}
			\ForEach{anomalous point}{
				prediction, score = $clf.predict(\text{$anomalous \ point$})$\;
				\If{prediction is non-anomalous}{
					Change point label assignment to match the current cluster\;
				}
				\Else{
					Leave anomaly label unchanged\;
				}
			}
		}
		$counter = counter + 1$\;
	}

	Re-fit the Perception() model on each cluster and compute scores\;
	\Return labels $L$ and membership scores $M$ of all examples\;
\end{algorithm}

\section{Experimental Results}
\label{section:example_results}
This section presents the evaluation results of the proposed clustering algorithm across a variety of datasets, including one-dimensional and two-dimensional synthetic datasets, as well as several publicly available benchmark datasets. We compare our method against a selection of representative clustering algorithms: $k$-means, DBSCAN, Agglomerative Clustering, Gaussian Mixture Models (GMMs), Seeded $k$-means (S-KM), Constrained $k$-means (C-KM), COP-KMeans (COPKM), and Deep Embedded Clustering (DEC). These methods were chosen because they are well-established baselines in the domains of unsupervised, semi-supervised, and deep clustering, and because reliable, well-written implementations are available. Implementation of $k$-means, Seeded $k$-means, DBSCAN, Agglomerative Clustering, and GMMs were obtained or adapted from the scikit-learn library~\cite{scikitlearn}. For Constrained $k$-means and COP-KMeans, we used open-source implementations from the Python packages described in~\cite{Levy-Kramer_k-means-constrained_2018, behrouz_babaki_2017_831850}. The DEC model was implemented using the publicly available ClustPy package~\cite{leiber2023benchmarking}. For all methods that require the number of ground truth clusters as input, we provided it explicitly. Although this could be seen as introducing a semi-supervised element, our intention was to ensure that algorithms like $k$-means were given their best chance to perform optimally with respect to this critical parameter. All other algorithm parameters were left at their default settings, reflecting the challenges of parameter tuning and the sensitivity of some methods to such choices in the absence of labelled data. The only exception here was for DEC, where the number of pretrain and clustering epochs were set to $10$ due to high computation time even at these low settings. 

Evaluating clustering algorithms is inherently challenging due to the wide range of available metrics and the fact that many of these measures inherently favour certain types of algorithms, depending on the optimisation criteria they employ. For example, partitioning algorithms often produce results that differ significantly from those of clustering methods designed to detect anomalies—such as the approach proposed in this paper—and may appear to outperform them under conventional evaluation metrics. This is partly because many benchmark datasets assign all instances to a cluster, without explicitly labelling anomalous or outlying points. As a result, even the so-called ground truth may not accurately reflect the underlying structure of the data. For instance, in synthetic Gaussian mixtures, it is reasonable to consider some points as outliers relative to the core density of each cluster (see Figure~\ref{fig:2d_gaussian_zoomed}). Furthermore, it is important to recognise that a dataset can often be clustered in multiple meaningful ways, depending on the features or criteria used. Consequently, even when ground truth labels are available, a clustering algorithm may discover valid and insightful groupings that do not align with the predefined labels. It has thus been recognised that human (with domain expertise) evaluation would provide a more meaningful measure of performance; however, this is a difficult time and resource consuming method. In light of these considerations, this work first adopts a qualitative evaluation strategy to better understand the patterns uncovered by the algorithms, followed by a quantitative automatic analysis using established external validation metrics to benchmark their performance.

\subsection{One-dimensional Synthetic Data}
\label{section:1d_results}
\begin{figure}
	\includegraphics[width=\linewidth]{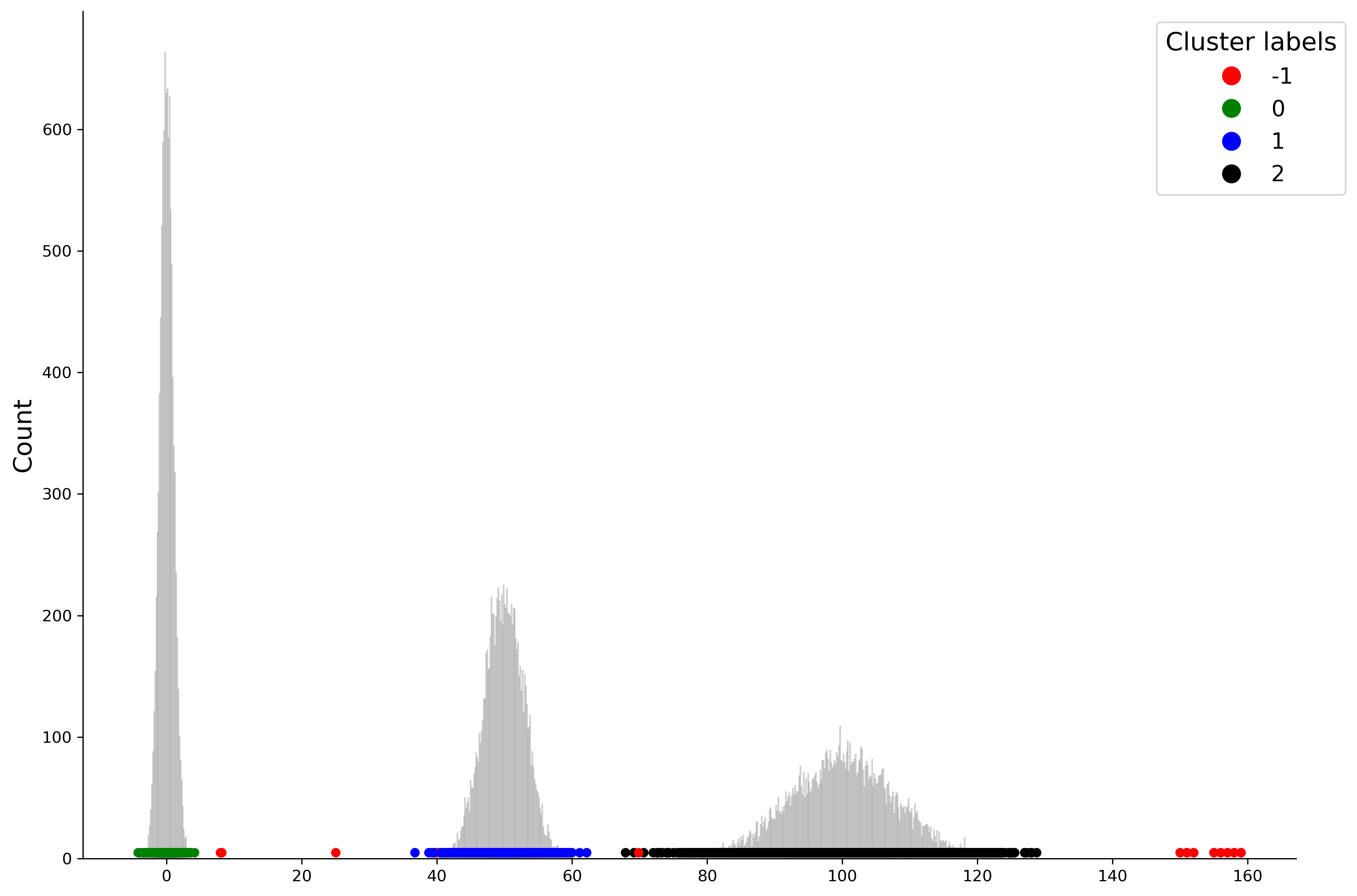}
	\caption{Three labelled Gaussian distributed clusters with a few additional mislabelled examples, some isolated anomalies and a small anomalous cluster.}
	\label{fig:1d_gaussian_y_true}
\end{figure}

The first dataset, named \textit{1d\_gauss}, consists of three Gaussian-distributed clusters, each of sizes $[10000, 5000, 2500]$ examples with centers $[0,50,100]$ and standard deviations of $[1,3,8]$. Added to this set are a few additional isolated local anomalies, a few mislabelled examples, and a small globally anomalous cluster. Figure \ref{fig:1d_gaussian_y_true} illustrates the data distribution where each of the three clusters and anomalies are labelled and coloured. Note however that anomalies resulting from the generating Gaussian distribution for each of the clusters are not highlighted as such since it would require selecting an anomaly detection algorithm which may give results different to another, and that such algorithms often require appropriate but unknown parameters. In addition to the labelled scatter plot, a histogram is also provided to show the relative densities which is otherwise hidden in the plot.

\begin{figure}
	\includegraphics[width=\linewidth]{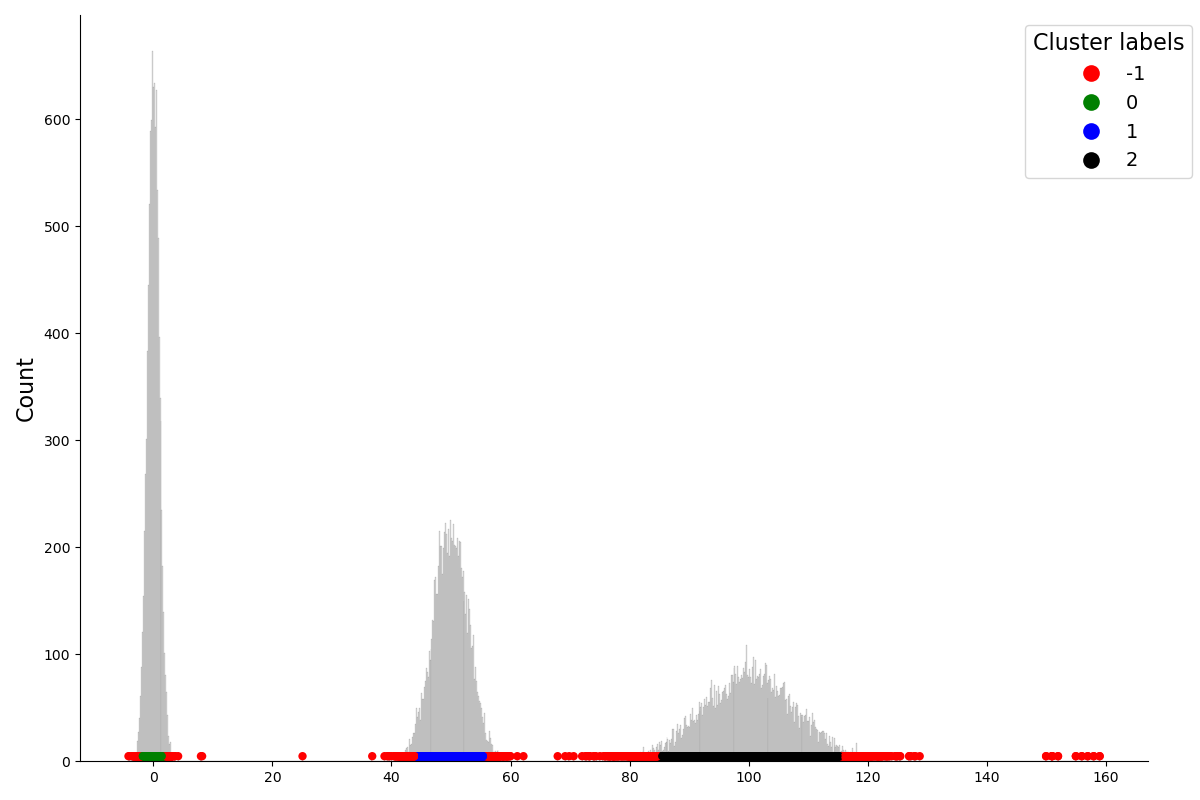}
	\caption{The proposed semi-supervised clustering algorithm successfully identifies the three underlying clusters while simultaneously detecting anomalies. Notably, the anomalous cluster is not forced into any of the labelled clusters, preserving the integrity of both the clustering and anomaly detection processes.}
	\label{fig:1d_nassir_results}
\end{figure}

\begin{figure}
	\includegraphics[width=\linewidth]{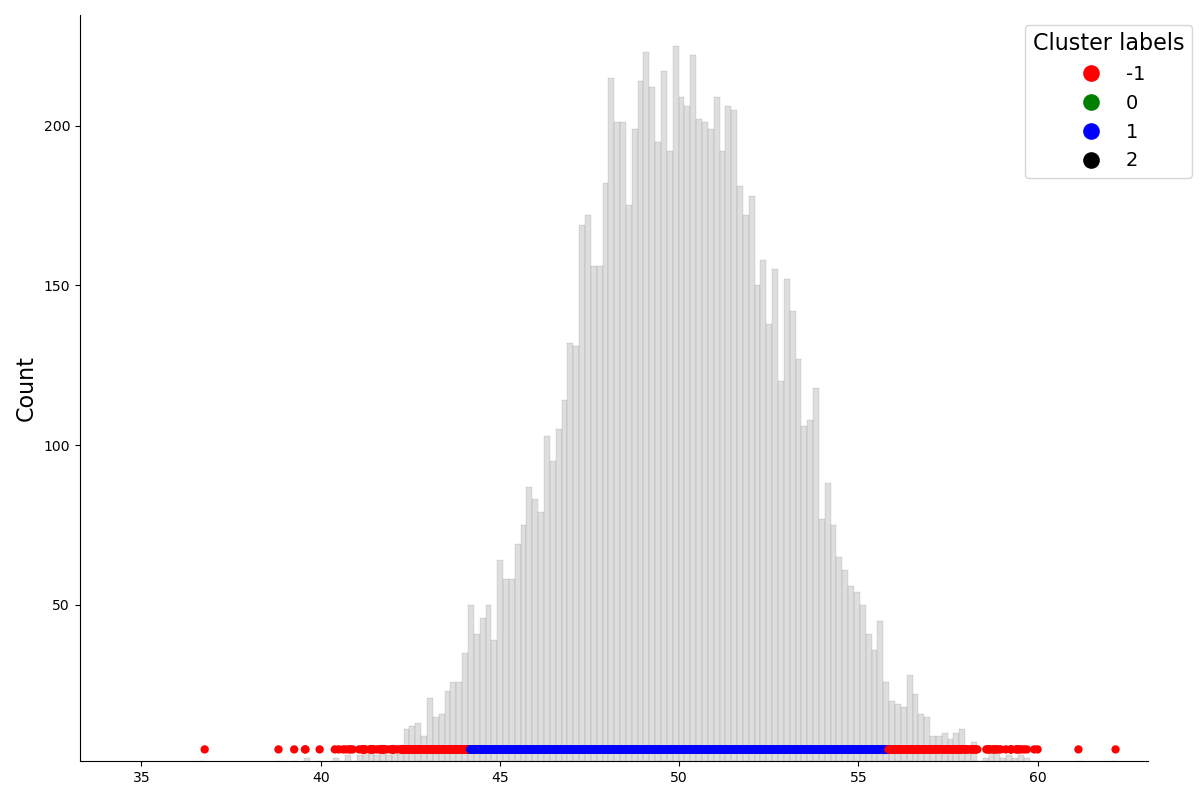}
	\caption{A close up view of cluster 1 from Figure \ref{fig:1d_nassir_results} showing the identification of not only anomalies and mislabelled data, but also fringe points that can be considered rare or unusual in a Gaussian distributed cluster of points.}
	\label{fig:1d_zoomed}
\end{figure}

\begin{figure}
	\includegraphics[width=\linewidth]{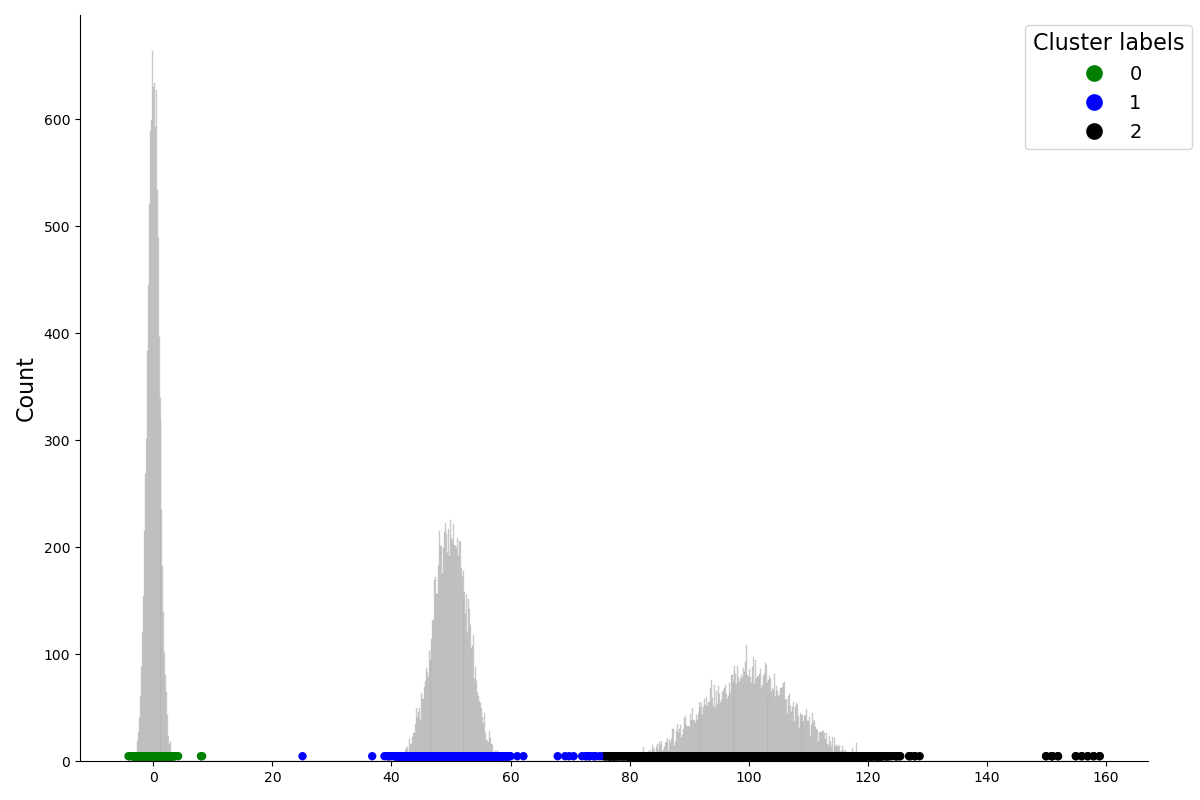}
	\caption{The result of applying $k$-means where all points are put into one of three groups. It has no notion of outliers in the data that do not belong to any cluster.}
	\label{fig:1d_kmeans_results}
\end{figure}

\begin{figure}
	\includegraphics[width=\linewidth]{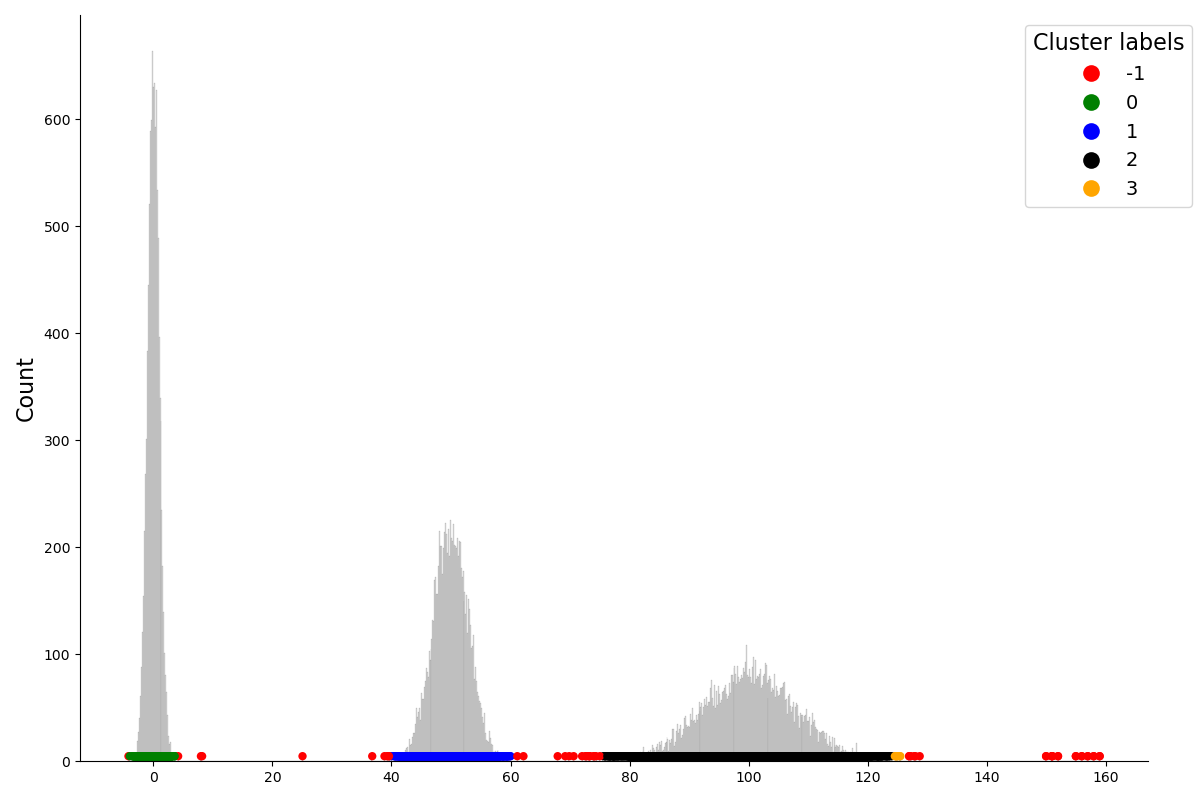}
	\caption{DBSCAN results using its default parameters are shown, where it largely detects the three clusters and also produces anomaly labels for points that were deemed insufficient to be part of a cluster.}
	\label{fig:1d_DBSCAN}
\end{figure}

A tiny fraction of the data---$35$ examples ($0.2\%$) together with their labels---are randomly chosen from the data (excluding the small anomalous cluster). These labels act as the seeds for the proposed semi-supervised clustering algorithm and are expected to have been provided by the user after familiarising themselves and interacting with the data and domain. Note that the seeds need not cover all clusters since unrecognised clusters will simply be reported as anomalies and can be analysed in subsequent rounds. The results of the algorithm are shown in Figure \ref{fig:1d_nassir_results} where three clusters are identified along with the isolated anomalies and small anomalous cluster; the latter of which can be investigated by the user in order to decide if it satsifies her criteria of being a new cluster. Note that the algorithm importantly flags the points in the outer regions---which we name fringe points---of each cluster as anomalous. Closer inspection of one of the clusters and its histogram distribution illustrates how this can be desirous since these points are strong contenders for being considered outliers compared to the main mass of the cluster (Figure \ref{fig:1d_zoomed}). Thus, the proposed method carries out a clustering of data as opposed to partitioning \emph{all} data points into some group. Indeed, $k$-means partitions all data into one of three clusters---including the anomalies (Figure \ref{fig:1d_kmeans_results}). DBSCAN, on the other hand produces arguably good results here as it largely detects the three clusters and also produces anomaly labels for points that were deemed insufficient to be part of a cluster (Figure \ref{fig:1d_DBSCAN}).
 
\subsection{Two-dimensional Synthetic Data}
\begin{figure}
	\includegraphics[width=\linewidth]{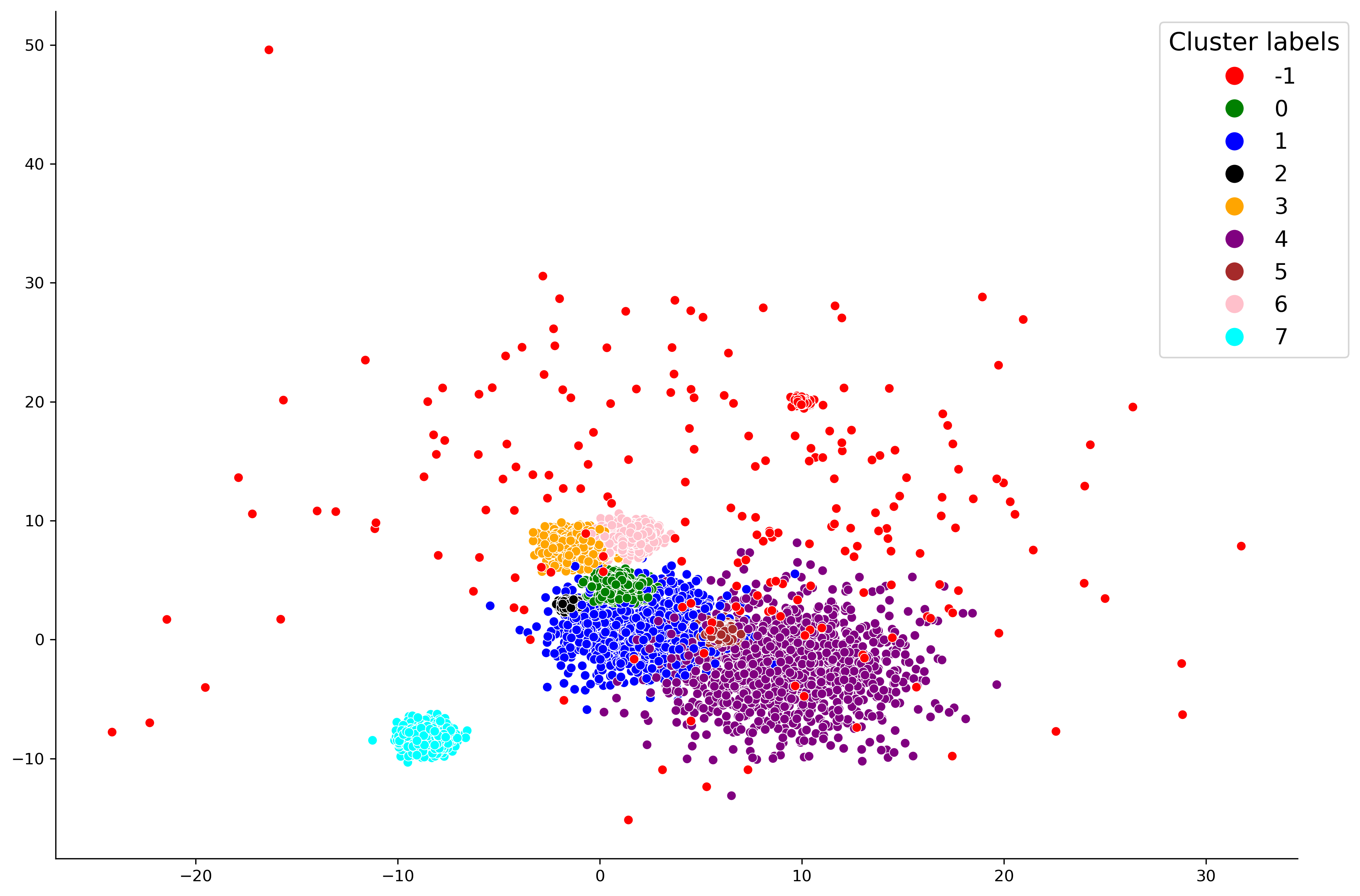}
	\caption{Eight Gaussian clusters are shown with differing standard deviations and locations. Added to the data is a scattering of anomalies and also a small anomalous cluster at coordinates $(11,20)$.}
	\label{fig:2d_gaussian}
\end{figure}

\begin{figure}
	\includegraphics[width=\linewidth]{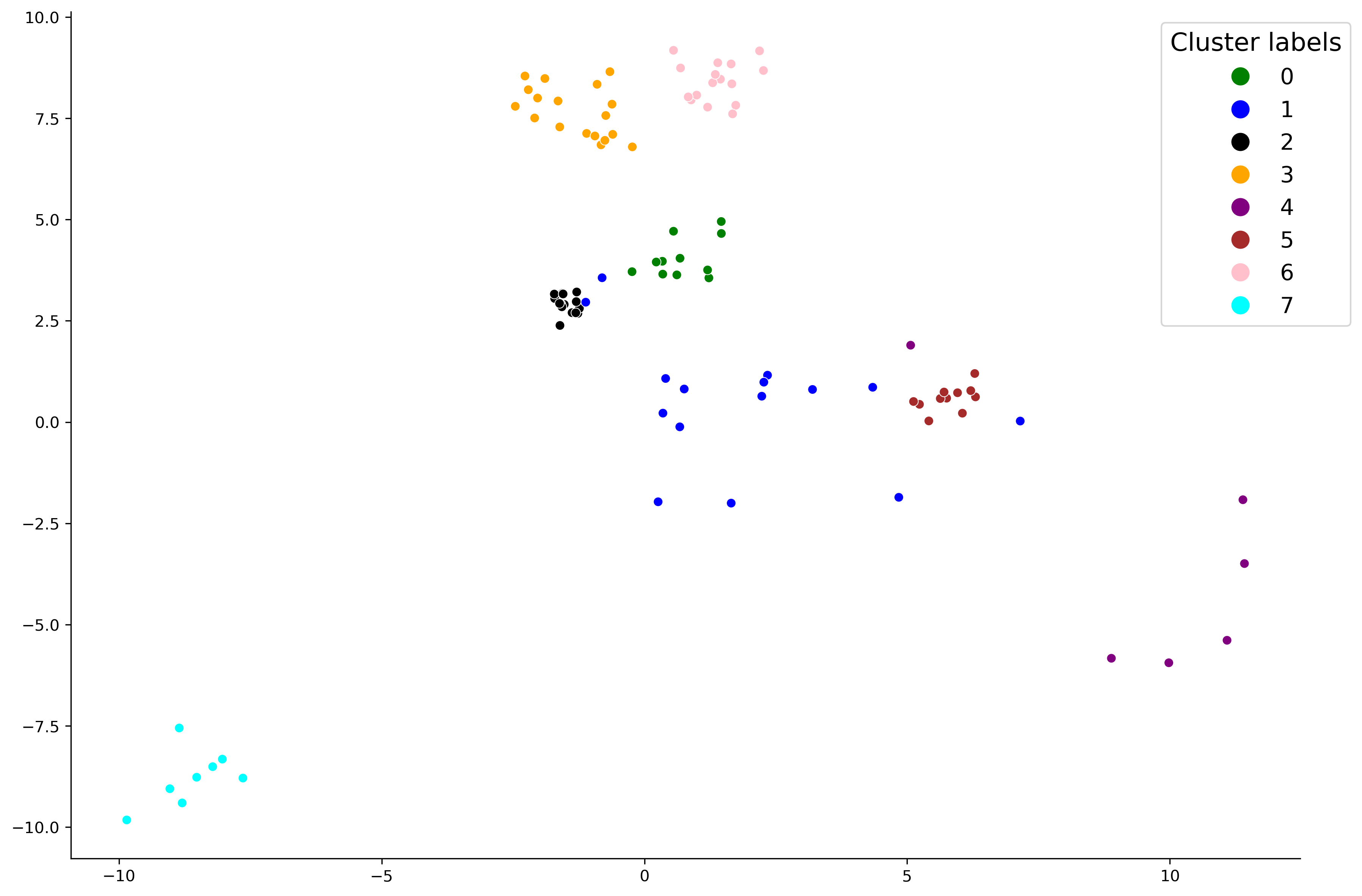}
	\caption{A random sample of points (1\% of the data) is shown from the Gaussian distributed clusters. These form the seeds of the proposed clustering algorithm.}
	\label{fig:2d_gaussian_seeds_only}
\end{figure}

The next example is a two-dimensional dataset, named \textit{2d\_gauss}, composed of eight random Gaussian distributed clusters at different locations and with standard deviations $[0.6, 2, 0.2, 0.7, 3, 0.4, 0.6, 0.6]$ (Figure \ref{fig:2d_gaussian}). Added to this are many isolated anomalies, and a relatively small anomalous cluster which could also be considered a newly emerging cluster. The total number of points is $10300$, and a small random sample of approximately $100$ points from the eight clusters are taken as seeds which is illustrated by Figure \ref{fig:2d_gaussian_seeds_only}. The number of seeds selected is such that we have at approximately $10$-$30$ per cluster in order to reliably build out the clusters. These would be expected to be provided by a domain expert and help guide the clustering algorithm to desired solutions. Note that the labels can even be noisy since mislabelled examples will be ejected out of the group as anomalies during algorithm execution.

\begin{figure}
	\includegraphics[width=\linewidth]{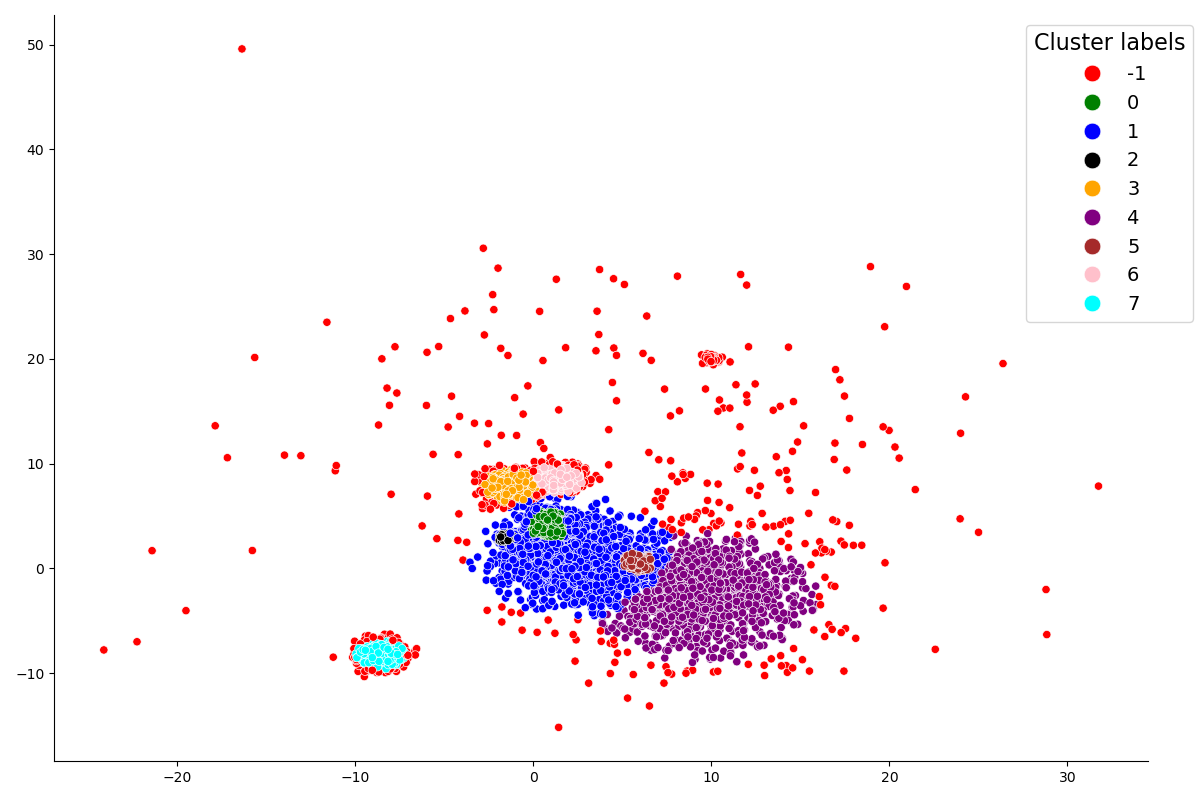}
	\caption{Results of applying the proposed clustering algorithm. The known clusters are largely recovered while points deemed anomalous with respect to the clusters are highlighted.}
	\label{fig:2d_gaussian_nassir_results}
\end{figure}

\begin{figure}
	\includegraphics[width=\linewidth]{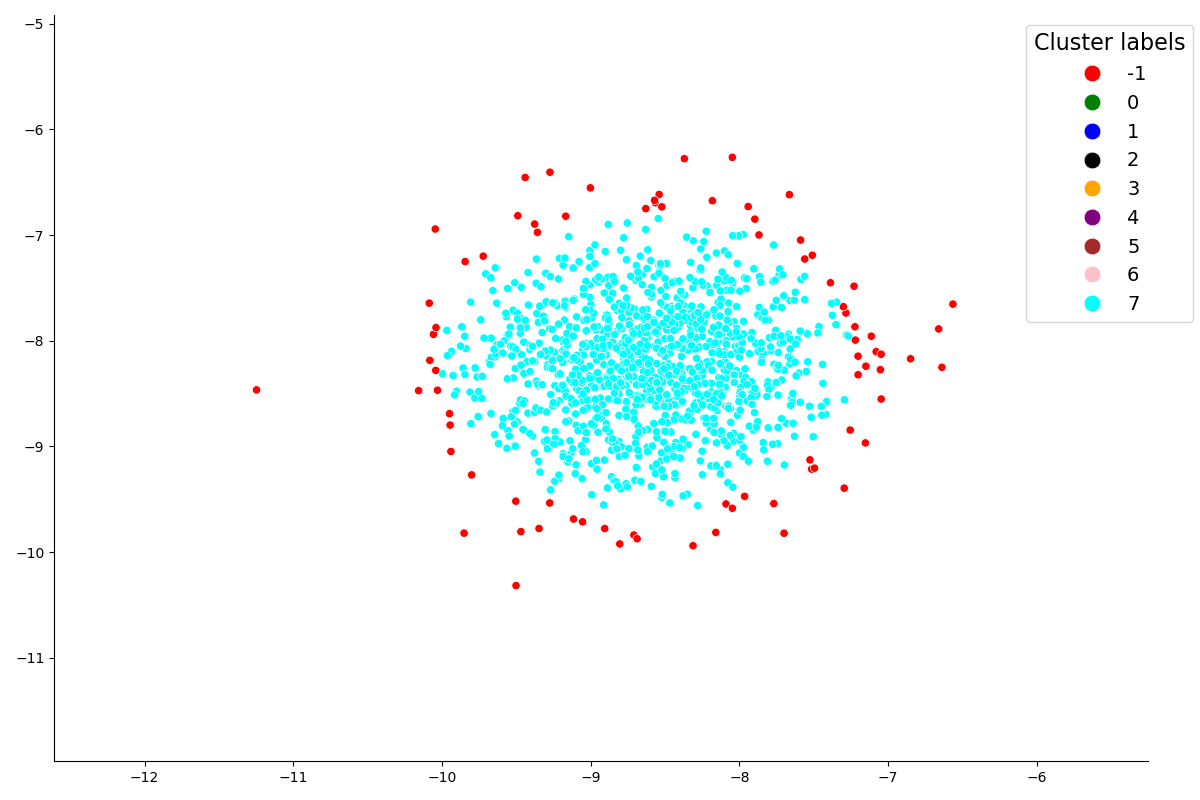}
	\caption{A close up view of cluster $7$ shows how the proposed clustering algorithm not only detects clusters, but also highlights anomalous and fringe points.}
	\label{fig:2d_gaussian_zoomed}
\end{figure}

\begin{figure}
	\includegraphics[width=\linewidth]{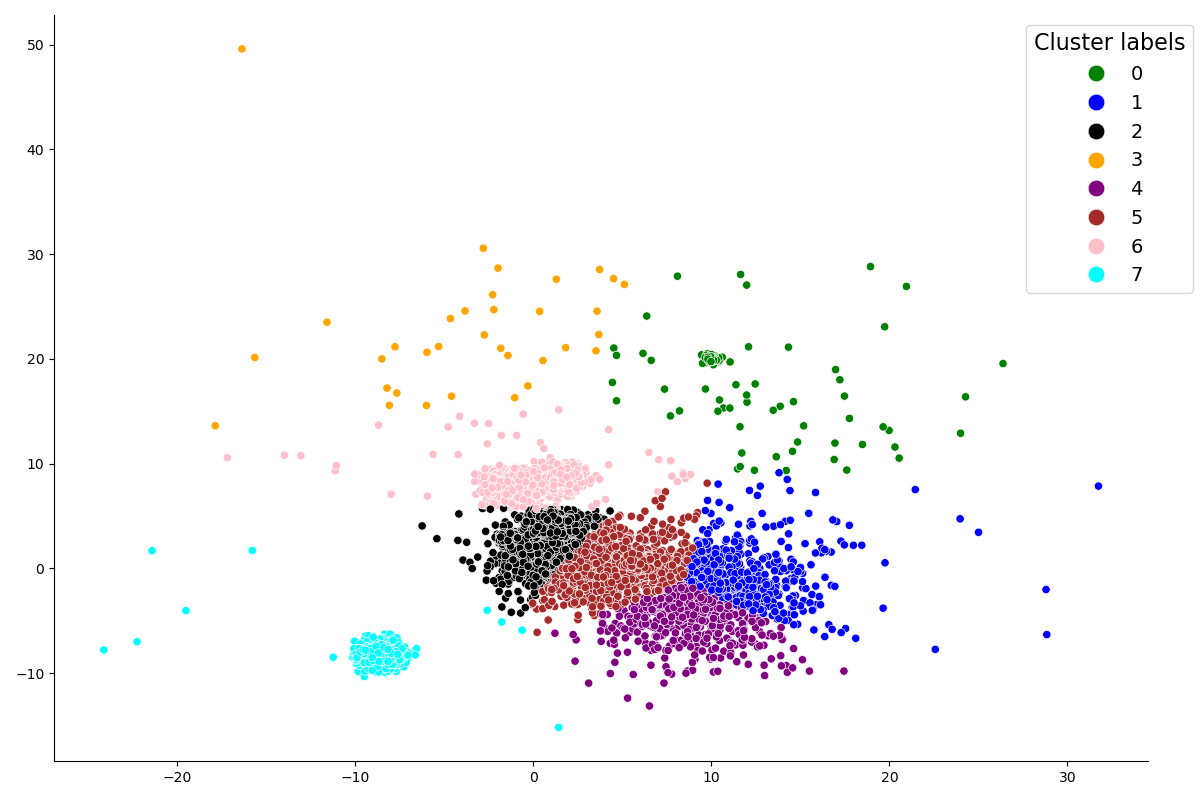}
	\caption{Results of $k$-means algorithm ($k=8$). Noteable observations are that some clusters are merged, data is unintuitively partioned and that anomalies are always assigned to some cluster.}
	\label{fig:2d_gaussian_kmeans}
\end{figure}

\begin{figure}
	\includegraphics[width=\linewidth]{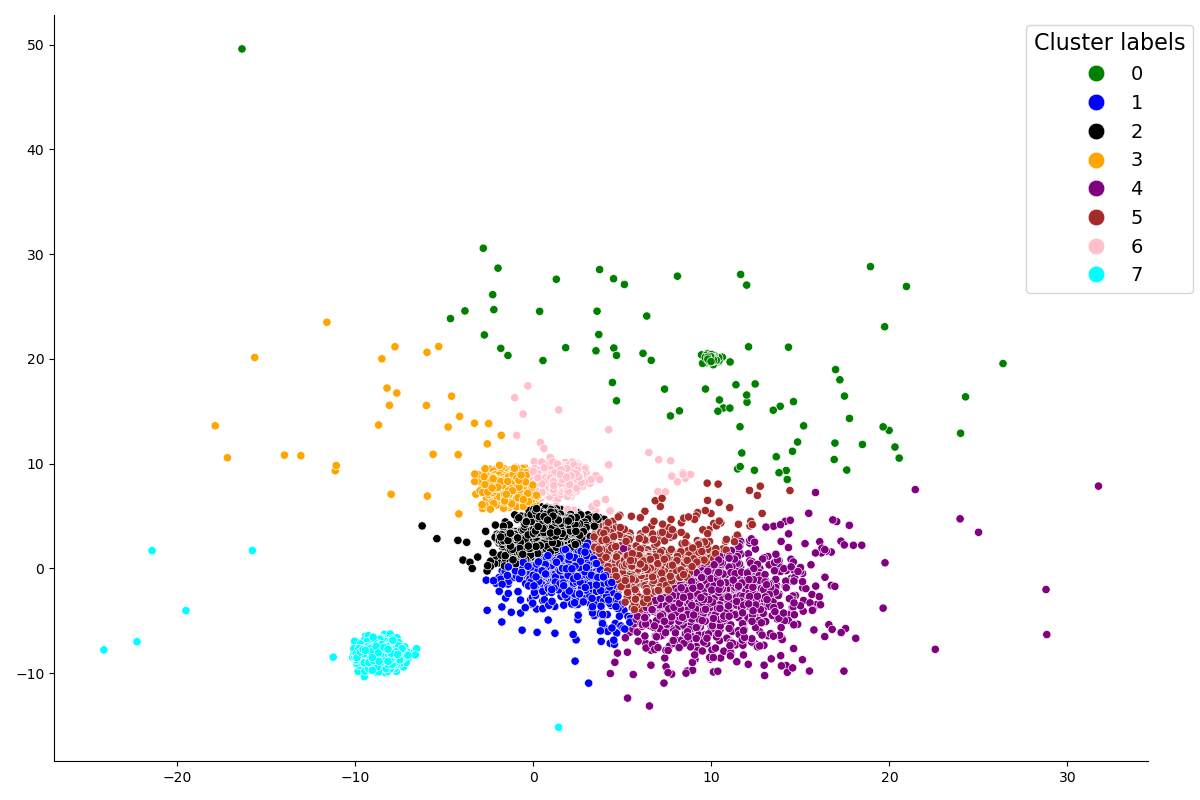}
	\caption{Results of applying the COP-KMeans algorithm, where even though the labelled data is used to form constraints, it still results in many undesired partitions of the data that are similar to $k$-means.}
	\label{fig:COPkmeans}
\end{figure}

The clustering result is shown in Figure \ref{fig:2d_gaussian_nassir_results} where the eight clusters are largely identified, some points around the clusters are deemed anomalous, and many isolated points and the small cluster labelled as anomalies. Note the difficulty of the task in that some clusters are close or overlapping, while others are located inside another. Hence, there are arguably misclassifications along bordering points. Figure \ref{fig:2d_gaussian_zoomed} also shows a close up view of one cluster to show the proposed methods ability to highlight fringe and far lying points as outliers with respect to the cluster's mass of points. Contrasting the results with that of $k$-means ($k=8$), we see in Figure \ref{fig:2d_gaussian_kmeans} that it unintuitvely partitions the points and does not reflect the desired groupings. Clusters originally labelled 3 and 6 are also grouped together and the small anomalous group of points is deemed to be a cluster. $k$-means also partitions all the data points resulting in anomalies being assigned to some cluster. The results obtained with DBSCAN exhibit interesting characteristics. The algorithm successfully identifies and labels many isolated points as outliers, accurately capturing their anomalous nature. It also considers the small group of points to be an anomalous cluster. However, it merges the majority of points into one large cluster, leading to a loss of granularity in the clustering results. It also does not report many of the points around the Gaussian clusters as anomalies (as in Figure \ref{fig:2d_gaussian_zoomed}). This behavior aligns with the algorithm's primary objective being to cluster data rather than detect anomalies. Constrained $k$-means and COP-KMeans also produce results similar to $k$-means, even though constraints are applied using the seed data; the result of applying COP-KMeans is shown in Figure \ref{fig:COPkmeans}. (We encourage readers to view results of other widely used clustering algorithms in the associated GitHub repository referenced on the first page).

\subsection{Multi-dimensional Realworld Examples}
This section presents the proposed clustering algorithm results on a selection of widely used benchmark datasets, alongside the synthetic datasets introduced in the previous section. These datasets were selected for their popularity, varied structural characteristics and differing levels of complexity~\cite{GonzalezAlmagroPPCG25}. All datasets were obtained from the UCI library of machine learning datasets~\cite{kelly2025uci} and the scikit-learn library~\cite{scikitlearn}. Evaluation was conducted quantitatively using a range of standard external validation metrics---Purity, V--Measure, Normalised Mutual Information (NMI), Adjusted Rand Index (ARI) and the Fowlkes–Mallows Index (FMI)—--to provide a comprehensive assessment of clustering quality. To focus solely on clustering performance, anomalous points were excluded from these analyses.

Table~\ref{tab:datasets} summarises the main characteristics of the datasets, showing the number of examples (n), features (Dim), and classes. In the \textit{yeast} and \textit{shuttle} datasets, minority classes were treated as anomalies due to their incoherent structure and resemblance to noise. These were reduced from $10$ and $7$ classes to $4$ and $3$, respectively. For datasets with more than $30$ features, dimensionality was reduced to $10$ using UMAP~\cite{mcinnes2018umap}, following a rule-of-thumb heuristic. This included \textit{ionosphere} ($34$) and \textit{MNIST digits} ($64$). An exception was made for \textit{cover\_type}, which was kept in its original high-dimensional form due to computational limitations, but as a consequence serves as a representative large-scale example.
In the case of the \textit{Newsgroups} dataset, which consists of text data, the feature count reflects the dimensionality after lemmatisation, TF-IDF vectorisation, and UMAP reduction. The table also states the percentage of labelled data used for semi-supervised methods; smaller datasets generally required a higher proportion of labels to guide clustering effectively as indicated by the average number of labels per cluster column. All datasets are available in the accompanying GitHub repository. Experiments were run on a MacBook M1 Pro with 16~GB of RAM.
\begin{table}[htbp]
\centering
\caption{Summary of benchmark datasets used in the evaluation.}
\label{tab:datasets}
\resizebox{\linewidth}{!}{%
\begin{tabular}{lrrrrr}
\toprule
\textbf{Dataset} & \textbf{n} & \textbf{Dim} & \textbf{\#Classes} & \textbf{\% Labelled} & \textbf{\#Lbl/Clust.} \\
\midrule
1d\_gauss           & 17,528   & 1   & 3   & 0.2  & 12  \\
2d\_gauss           & 10,300   & 2   & 8   & 1    & 13  \\
6Newsgroups\_UMAP10 & 10,496   & 10  & 6   & 1    & 17  \\
MNIST\_UMAP10       & 1,797    & 10  & 10  & 5    & 9   \\
banknote            & 1,372    & 4   & 2   & 2    & 13  \\
breast\_cancer      & 569      & 30  & 2   & 7    & 20  \\
cover\_type         & 581,012  & 54  & 7   & 0.02 & 15 \\
glass               & 214      & 9   & 6   & 30   & 10  \\
ionosphere\_UMAP10  & 351      & 10  & 2   & 10   & 17  \\
iris                & 150      & 4   & 3   & 20   & 10  \\
land\_mines         & 338      & 3   & 5   & 30   & 20  \\
pendigits           & 10,992   & 16  & 10  & 2.5  & 27  \\
seeds               & 210      & 7   & 3   & 20   & 14  \\
shuttle             & 43,500   & 9   & 3   & 0.2  & 29  \\
wine                & 178      & 13  & 3   & 30   & 17  \\
yeast               & 1,484    & 8   & 4  & 5     & 18   \\
\bottomrule
\end{tabular}
}
\end{table}

\begin{figure}[htbp]
    \centering
    \includegraphics[width=\textwidth]{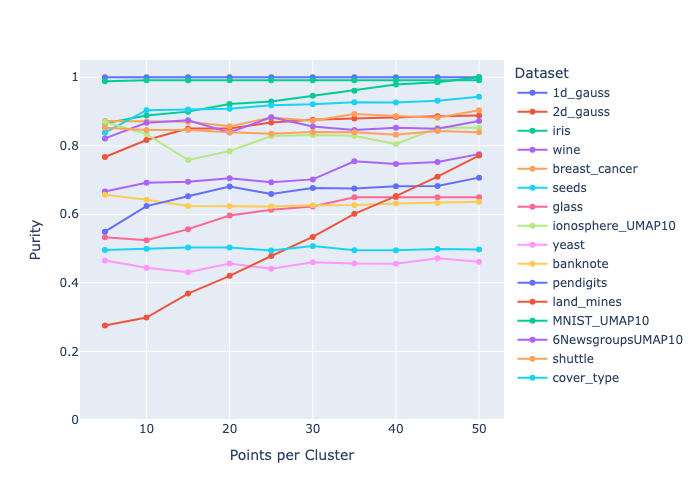}
    \caption{Purity scores of the proposed method as a function of the number of points per cluster.}
    \label{fig:points_per_cluster_purity}
\end{figure}

\begin{figure}[htbp]
    \centering
    \includegraphics[width=\textwidth]{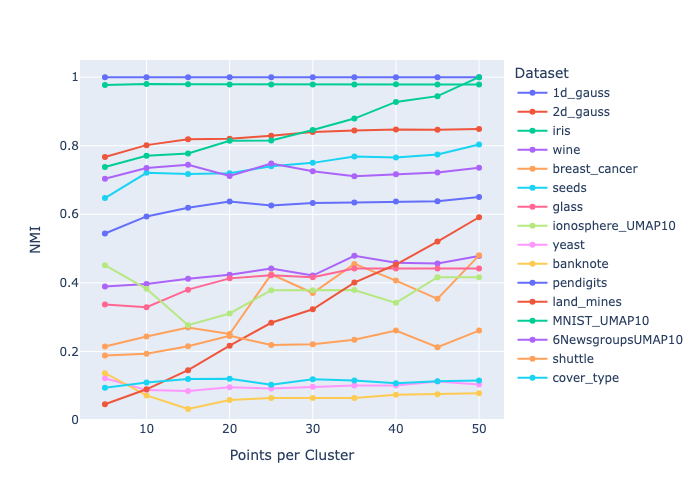}
    \caption{Normalized Mutual Information (NMI) scores of the proposed method as a function of the number of points per cluster.}
    \label{fig:nmi_vs_points_per_cluster}
\end{figure}

Figures \ref{fig:points_per_cluster_purity} and \ref{fig:nmi_vs_points_per_cluster} plot the purity and NMI performance of the proposed method against the number of seeds per cluster at initialisation across the datasets averaged over ten runs. They indicate that rather than the number of seeds required to be a percentage of the dataset, generally $10-30$ seeds per cluster are sufficient for effective clustering.

Tables~\ref{tab:purity}–\ref{tab:runtime} present the performance of the proposed semi-supervised clustering method across a diverse suite of benchmark datasets, in comparison with several established unsupervised and semi-supervised approaches. The method consistently delivers strong results across synthetic, real-world, and high-dimensional settings. It ranks among the top performers on structured synthetic datasets such as \textit{1d\_gauss} and \textit{2d\_gauss}, validating its correctness in controlled conditions. On more complex benchmarks—such as \textit{MNIST\_UMAP10} and \textit{6Newsgroups\_UMAP10}—it achieves the highest scores across all evaluation metrics, outperforming traditional, density-based, and deep clustering baselines, including DEC. While its performance is somewhat lower on certain datasets such as \textit{iris}, \textit{seeds}, and \textit{wine}, where methods like GMM tend to excel, the proposed method remains highly competitive overall. As expected, no single method performs best across all datasets and metrics, but the proposed approach demonstrates a favourable balance of robustness and versatility. (Dashed lines indicate instances where computational constraints prevented clustering.)

Importantly, the gains in clustering quality are achieved with competitive computational cost and it was able to cluster all datasets. It runs efficiently on most datasets, requiring no parameter tuning and only a small fraction of labelled data—underscoring its robustness and ease of deployment. Although runtimes are relatively higher on datasets such as \textit{cover\_type}, \textit{MNIST\_UMAP10}, and \textit{pendigits} compared to lightweight methods like $k$-means, profiling reveals that the overhead arises primarily from redundant cluster expansion steps and preprocessing in the underlying anomaly detection model. These findings point to clear opportunities for optimisation—both at the algorithmic level and through improvements in code efficiency or implementation language.


In summary, the proposed method strikes a strong balance between accuracy, efficiency, and scalability. Its ability to work with minimal supervision, reject noisy seeds during cluster expansion, and leave outliers or unknown groups unassigned for subsequent analysis further underscores its robustness and versatility—making it a practical and competitive choice for semi-supervised clustering in both research and applied contexts.

\begin{table}
  \centering
  \resizebox{\linewidth}{!}{%
    \begin{tabular}{lrrrrrrrrr}
\toprule
 & KMeans & GMM & S-KM & C-KM & COPKM & Agg & DBSCAN & DEC & Proposed \\
dataset &  &  &  &  &  &  &  &  &  \\
\midrule
1d\_gauss & 1.00 & 1.00 & 1.00 & 0.93 & 1.00 & 1.00 & 1.00 & 1.00 & 1.00 \\
2d\_gauss & 0.81 & 0.88 & 0.84 & 0.81 & 0.77 & 0.78 & 0.52 & 0.84 & 0.89 \\
6NewsgroupsUMAP10 & 0.74 & 0.74 & 0.78 & 0.74 & 0.74 & 0.72 & 0.34 & 0.74 & 0.89 \\
MNIST\_UMAP10 & 0.88 & 0.88 & 0.88 & 0.87 & 0.87 & 0.88 & 0.97 & 0.88 & 0.99 \\
banknote & 0.61 & 0.56 & 0.61 & 0.61 & 0.62 & 0.56 & 0.70 & 0.56 & 0.62 \\
breast\_cancer & 0.85 & 0.95 & 0.85 & 0.87 & 0.86 & 0.78 & 0.63 & 0.85 & 0.90 \\
cover\_type & 0.49 & 0.57 & 0.49 & -- & -- & -- & -- & 0.49 & 0.50 \\
glass & 0.59 & 0.56 & 0.57 & 0.62 & 0.62 & 0.54 & 0.49 & 0.43 & 0.56 \\
ionosphere\_UMAP10 & 0.70 & 0.69 & 0.70 & 0.70 & 0.72 & 0.70 & 0.64 & 0.68 & 0.85 \\
iris & 0.89 & 0.97 & 0.89 & 0.89 & 0.79 & 0.89 & 0.69 & 0.93 & 0.89 \\
land\_mines & 0.23 & 0.38 & 0.30 & 0.30 & 0.45 & 0.28 & 0.21 & 0.26 & 0.41 \\
pendigits & 0.75 & 0.71 & 0.77 & 0.77 & 0.71 & 0.72 & 0.10 & 0.75 & 0.69 \\
seeds & 0.90 & 0.93 & 0.90 & 0.90 & 0.91 & 0.89 & 0.53 & 0.86 & 0.90 \\
shuttle & 0.79 & 0.79 & 0.83 & 0.79 & -- & -- & 0.79 & 0.79 & 0.84 \\
wine & 0.70 & 0.85 & 0.70 & 0.70 & 0.78 & 0.70 & 0.40 & 0.70 & 0.70 \\
yeast & 0.49 & 0.44 & 0.50 & 0.49 & 0.49 & 0.47 & 0.36 & 0.47 & 0.44 \\
\bottomrule
\end{tabular}

  }
  \caption{Purity metric results}
  \label{tab:purity}
\end{table}

\begin{table}
  \centering
  \resizebox{\linewidth}{!}{%
    \begin{tabular}{lrrrrrrrrr}
\toprule
 & KMeans & GMM & S-KM & C-KM & COPKM & Agg & DBSCAN & DEC & Proposed \\
dataset &  &  &  &  &  &  &  &  &  \\
\midrule
1d\_gauss & 1.00 & 1.00 & 1.00 & 0.80 & 1.00 & 1.00 & 1.00 & 1.00 & 1.00 \\
2d\_gauss & 0.87 & 0.92 & 0.83 & 0.83 & 0.82 & 0.85 & 0.72 & 0.85 & 0.84 \\
6NewsgroupsUMAP10 & 0.61 & 0.61 & 0.62 & 0.60 & 0.60 & 0.61 & 0.29 & 0.62 & 0.76 \\
MNIST\_UMAP10 & 0.90 & 0.90 & 0.90 & 0.87 & 0.89 & 0.90 & 0.93 & 0.90 & 0.98 \\
banknote & 0.03 & 0.01 & 0.03 & 0.03 & 0.03 & 0.00 & 0.17 & 0.01 & 0.01 \\
breast\_cancer & 0.46 & 0.71 & 0.46 & 0.50 & 0.48 & 0.32 & 0.00 & 0.46 & 0.49 \\
cover\_type & 0.07 & 0.17 & 0.07 & -- & -- & -- & -- & 0.04 & 0.11 \\
glass & 0.43 & 0.40 & 0.40 & 0.36 & 0.45 & 0.39 & 0.31 & 0.14 & 0.36 \\
ionosphere\_UMAP10 & 0.12 & 0.11 & 0.12 & 0.12 & 0.16 & 0.12 & 0.10 & 0.11 & 0.42 \\
iris & 0.74 & 0.90 & 0.74 & 0.76 & 0.63 & 0.77 & 0.60 & 0.81 & 0.77 \\
land\_mines & 0.01 & 0.20 & 0.10 & 0.12 & 0.18 & 0.08 & 0.00 & 0.01 & 0.20 \\
pendigits & 0.69 & 0.68 & 0.69 & 0.69 & 0.68 & 0.73 & 0.00 & 0.74 & 0.64 \\
seeds & 0.69 & 0.76 & 0.69 & 0.69 & 0.73 & 0.73 & 0.29 & 0.65 & 0.69 \\
shuttle & 0.00 & 0.00 & 0.26 & 0.16 & -- & -- & 0.00 & 0.08 & 0.21 \\
wine & 0.43 & 0.58 & 0.43 & 0.43 & 0.50 & 0.42 & 0.00 & 0.43 & 0.41 \\
yeast & 0.12 & 0.11 & 0.16 & 0.12 & 0.13 & 0.13 & 0.00 & 0.09 & 0.08 \\
\bottomrule
\end{tabular}

  }
  \caption{V-Measure metric results}
  \label{tab:vmeasure}
\end{table}

\begin{table}
  \centering
  \resizebox{\linewidth}{!}{%
    \begin{tabular}{lrrrrrrrrr}
\toprule
 & KMeans & GMM & S-KM & C-KM & COPKM & Agg & DBSCAN & DEC & Proposed \\
dataset &  &  &  &  &  &  &  &  &  \\
\midrule
1d\_gauss & 1.00 & 1.00 & 1.00 & 0.80 & 1.00 & 1.00 & 1.00 & 1.00 & 1.00 \\
2d\_gauss & 0.87 & 0.92 & 0.83 & 0.83 & 0.82 & 0.85 & 0.72 & 0.85 & 0.84 \\
6NewsgroupsUMAP10 & 0.61 & 0.61 & 0.62 & 0.60 & 0.60 & 0.61 & 0.29 & 0.62 & 0.76 \\
MNIST\_UMAP10 & 0.90 & 0.90 & 0.90 & 0.87 & 0.89 & 0.90 & 0.93 & 0.90 & 0.98 \\
banknote & 0.03 & 0.01 & 0.03 & 0.03 & 0.03 & 0.00 & 0.17 & 0.01 & 0.01 \\
breast\_cancer & 0.46 & 0.71 & 0.46 & 0.50 & 0.48 & 0.32 & 0.00 & 0.46 & 0.49 \\
cover\_type & 0.07 & 0.17 & 0.07 & -- & -- & -- & -- & 0.04 & 0.11 \\
glass & 0.43 & 0.40 & 0.40 & 0.36 & 0.45 & 0.39 & 0.31 & 0.14 & 0.36 \\
ionosphere\_UMAP10 & 0.12 & 0.11 & 0.12 & 0.12 & 0.16 & 0.12 & 0.10 & 0.11 & 0.42 \\
iris & 0.74 & 0.90 & 0.74 & 0.76 & 0.63 & 0.77 & 0.60 & 0.81 & 0.77 \\
land\_mines & 0.01 & 0.20 & 0.10 & 0.12 & 0.18 & 0.08 & 0.00 & 0.01 & 0.20 \\
pendigits & 0.69 & 0.68 & 0.69 & 0.69 & 0.68 & 0.73 & 0.00 & 0.74 & 0.64 \\
seeds & 0.69 & 0.76 & 0.69 & 0.69 & 0.73 & 0.73 & 0.29 & 0.65 & 0.69 \\
shuttle & 0.00 & 0.00 & 0.26 & 0.16 & -- & -- & 0.00 & 0.08 & 0.21 \\
wine & 0.43 & 0.58 & 0.43 & 0.43 & 0.50 & 0.42 & 0.00 & 0.43 & 0.41 \\
yeast & 0.12 & 0.11 & 0.16 & 0.12 & 0.13 & 0.13 & 0.00 & 0.09 & 0.08 \\
\bottomrule
\end{tabular}

  }
  \caption{NMI metric results}
  \label{tab:nmi}
\end{table}

\begin{table}
  \centering
  \resizebox{\linewidth}{!}{%
    \begin{tabular}{lrrrrrrrrr}
\toprule
 & KMeans & GMM & S-KM & C-KM & COPKM & Agg & DBSCAN & DEC & Proposed \\
dataset &  &  &  &  &  &  &  &  &  \\
\midrule
1d\_gauss & 1.00 & 1.00 & 1.00 & 0.77 & 1.00 & 1.00 & 1.00 & 1.00 & 1.00 \\
2d\_gauss & 0.76 & 0.85 & 0.73 & 0.73 & 0.70 & 0.72 & 0.43 & 0.80 & 0.78 \\
6NewsgroupsUMAP10 & 0.61 & 0.60 & 0.63 & 0.60 & 0.58 & 0.55 & 0.12 & 0.60 & 0.77 \\
MNIST\_UMAP10 & 0.81 & 0.81 & 0.82 & 0.79 & 0.81 & 0.81 & 0.91 & 0.82 & 0.98 \\
banknote & 0.05 & 0.00 & 0.05 & 0.05 & 0.05 & 0.00 & 0.02 & 0.01 & 0.02 \\
breast\_cancer & 0.49 & 0.81 & 0.49 & 0.54 & 0.51 & 0.29 & 0.00 & 0.49 & 0.62 \\
cover\_type & -0.00 & 0.09 & -0.00 & -- & -- & -- & -- & 0.01 & 0.05 \\
glass & 0.27 & 0.25 & 0.27 & 0.22 & 0.27 & 0.26 & 0.16 & 0.06 & 0.18 \\
ionosphere\_UMAP10 & 0.15 & 0.14 & 0.15 & 0.15 & 0.20 & 0.15 & -0.07 & 0.13 & 0.49 \\
iris & 0.72 & 0.90 & 0.72 & 0.73 & 0.57 & 0.73 & 0.52 & 0.82 & 0.74 \\
land\_mines & -0.01 & 0.07 & 0.03 & 0.04 & 0.11 & 0.01 & 0.00 & 0.00 & 0.05 \\
pendigits & 0.58 & 0.54 & 0.60 & 0.60 & 0.56 & 0.55 & 0.00 & 0.60 & 0.52 \\
seeds & 0.72 & 0.80 & 0.72 & 0.72 & 0.76 & 0.71 & 0.05 & 0.64 & 0.71 \\
shuttle & 0.00 & 0.00 & 0.46 & 0.10 & -- & -- & 0.00 & 0.04 & 0.23 \\
wine & 0.37 & 0.61 & 0.37 & 0.37 & 0.48 & 0.37 & 0.00 & 0.37 & 0.36 \\
yeast & 0.09 & 0.08 & 0.11 & 0.10 & 0.10 & 0.06 & -0.00 & 0.07 & 0.06 \\
\bottomrule
\end{tabular}

  }
  \caption{ARI metric results}
  \label{tab:ari}
\end{table}

\begin{table}
  \centering
  \resizebox{\linewidth}{!}{%
    \begin{tabular}{lrrrrrrrrr}
\toprule
 & KMeans & GMM & S-KM & C-KM & COPKM & Agg & DBSCAN & DEC & Proposed \\
dataset &  &  &  &  &  &  &  &  &  \\
\midrule
1d\_gauss & 1.00 & 1.00 & 1.00 & 0.87 & 1.00 & 1.00 & 1.00 & 1.00 & 1.00 \\
2d\_gauss & 0.80 & 0.87 & 0.77 & 0.77 & 0.75 & 0.77 & 0.60 & 0.82 & 0.81 \\
6NewsgroupsUMAP10 & 0.68 & 0.68 & 0.70 & 0.67 & 0.66 & 0.64 & 0.44 & 0.68 & 0.81 \\
MNIST\_UMAP10 & 0.84 & 0.83 & 0.84 & 0.81 & 0.83 & 0.83 & 0.92 & 0.85 & 0.98 \\
banknote & 0.55 & 0.51 & 0.55 & 0.55 & 0.55 & 0.50 & 0.51 & 0.52 & 0.60 \\
breast\_cancer & 0.79 & 0.91 & 0.79 & 0.81 & 0.80 & 0.74 & 0.73 & 0.79 & 0.84 \\
cover\_type & 0.26 & 0.36 & 0.26 & -- & -- & -- & -- & 0.24 & 0.34 \\
glass & 0.51 & 0.48 & 0.49 & 0.40 & 0.45 & 0.51 & 0.43 & 0.32 & 0.40 \\
ionosphere\_UMAP10 & 0.59 & 0.58 & 0.59 & 0.59 & 0.61 & 0.59 & 0.54 & 0.58 & 0.77 \\
iris & 0.81 & 0.94 & 0.81 & 0.82 & 0.71 & 0.82 & 0.71 & 0.88 & 0.82 \\
land\_mines & 0.19 & 0.31 & 0.23 & 0.24 & 0.29 & 0.23 & 0.44 & 0.20 & 0.35 \\
pendigits & 0.62 & 0.59 & 0.64 & 0.64 & 0.61 & 0.61 & 0.32 & 0.65 & 0.57 \\
seeds & 0.81 & 0.87 & 0.81 & 0.81 & 0.84 & 0.81 & 0.47 & 0.76 & 0.81 \\
shuttle & 0.80 & 0.81 & 0.84 & 0.61 & -- & -- & 0.80 & 0.80 & 0.73 \\
wine & 0.58 & 0.74 & 0.58 & 0.58 & 0.66 & 0.58 & 0.58 & 0.58 & 0.59 \\
yeast & 0.39 & 0.45 & 0.46 & 0.39 & 0.39 & 0.47 & 0.53 & 0.34 & 0.37 \\
\bottomrule
\end{tabular}

  }
  \caption{FMI metric results}
  \label{tab:fmi}
\end{table}

\begin{table}
  \centering
  \resizebox{\linewidth}{!}{%
    \begin{tabular}{lrrrrrrrrr}
\toprule
 & KMeans & GMM & S-KM & C-KM & COPKM & Agg & DBSCAN & DEC & Proposed \\
dataset &  &  &  &  &  &  &  &  &  \\
\midrule
1d\_gauss & 0.05 & 0.04 & 0.01 & 1.50 & 3.59 & 4.81 & 0.22 & 17.92 & 0.12 \\
2d\_gauss & 0.02 & 0.14 & 0.01 & 5.89 & 4.80 & 1.40 & 0.07 & 12.17 & 0.11 \\
6NewsgroupsUMAP10 & 0.02 & 0.11 & 0.01 & 2.25 & 1.52 & 0.42 & 0.09 & 6.76 & 0.12 \\
MNIST\_UMAP10 & 0.01 & 0.03 & 0.00 & 0.48 & 0.47 & 0.03 & 0.01 & 2.58 & 2.60 \\
banknote & 0.01 & 0.01 & 0.00 & 0.20 & 0.08 & 0.02 & 0.00 & 2.04 & 0.02 \\
breast\_cancer & 0.01 & 0.10 & 0.00 & 0.05 & 0.12 & 0.00 & 0.01 & 0.85 & 0.01 \\
cover\_type & 1.04 & 81.25 & 0.99 & -- & -- & -- & -- & 570.80 & 26.27 \\
glass & 0.00 & 0.01 & 0.00 & 0.07 & 0.64 & 0.00 & 0.00 & 0.37 & 0.79 \\
ionosphere\_UMAP10 & 0.00 & 0.00 & 0.00 & 0.02 & 0.02 & 0.00 & 0.00 & 0.59 & 0.01 \\
iris & 0.01 & 0.01 & 0.00 & 0.03 & 0.01 & 0.00 & 0.00 & 0.32 & 0.00 \\
land\_mines & 0.01 & 0.03 & 0.00 & 0.11 & 0.17 & 0.00 & 0.00 & 0.58 & 0.70 \\
pendigits & 0.04 & 2.00 & 0.01 & 22.32 & 13.04 & 2.10 & 0.43 & 14.04 & 0.14 \\
seeds & 0.00 & 0.01 & 0.00 & 0.03 & 0.02 & 0.00 & 0.00 & 0.35 & 0.00 \\
shuttle & 0.17 & 0.52 & 0.03 & 8.16 & -- & -- & 0.55 & 46.73 & 0.77 \\
wine & 0.00 & 0.01 & 0.00 & 0.04 & 0.04 & 0.00 & 0.00 & 0.32 & 0.00 \\
yeast & 0.02 & 0.13 & 0.01 & 1.62 & 0.24 & 0.02 & 0.04 & 2.22 & 0.02 \\
\bottomrule
\end{tabular}

  }
  \caption{Runtime (s) metric results}
  \label{tab:runtime}
\end{table}

\subsubsection{MNIST Handwritten Digits}

\begin{figure}
	\includegraphics[width=\linewidth]{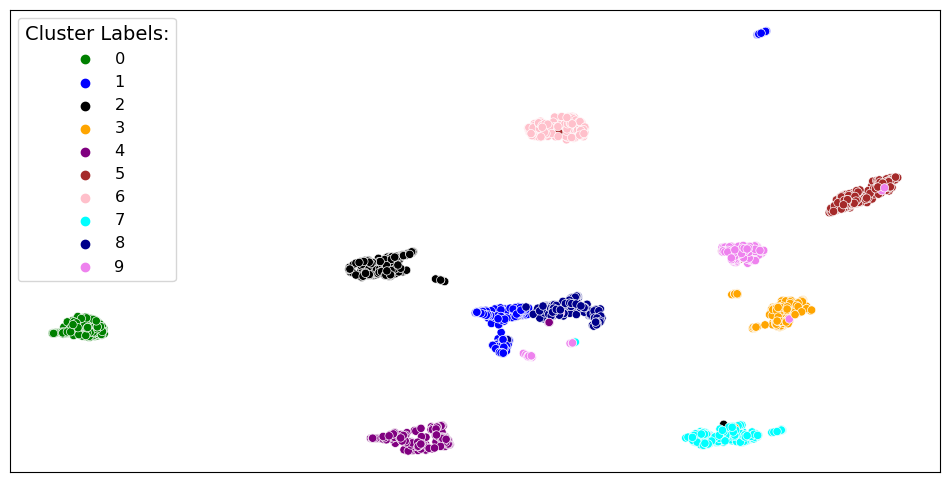}
	\caption{The 2-dimensional UMAP of the MNIST digits dataset obtained from scikit-learn, using the cosine metric. Each data point is color-coded based on its corresponding class label obtained from the provided labels.}
	\label{fig:digits_umap_labelled}
\end{figure}
\begin{figure}
	\includegraphics[width=\linewidth]{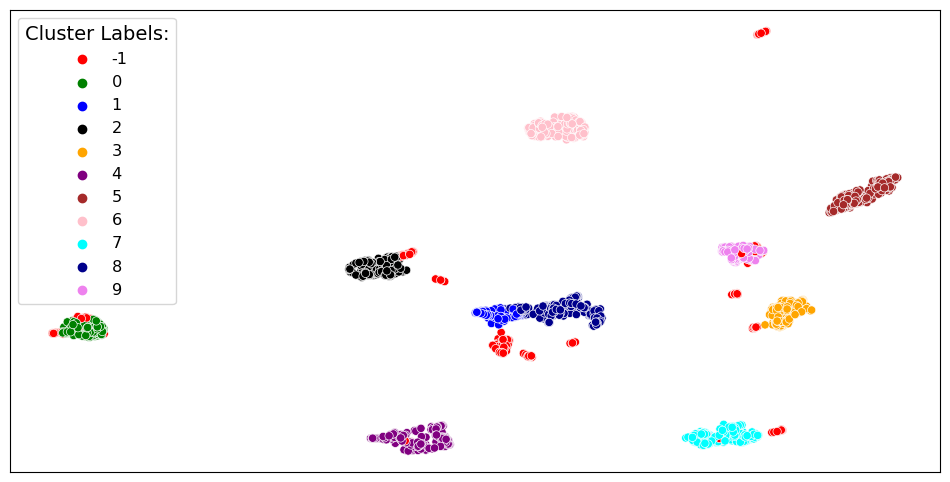}
	\caption{The results of applying the clustering algorithm to the MNIST digits data set obtained from sklearn after performing a UMAP transformation. The actual clustering process involved mapping the data to $10$ dimensions. However, for the purpose of visualisation, the results are presented in a 2-dimensional mapping.}
	\label{fig:digits_umap_nassir}
\end{figure}

This section presents the visual results of applying the proposed clustering algorithm to the MNIST digits dataset. Figure \ref{fig:digits_umap_labelled} depicts the application of UMAP to 2 dimensions using the cosine metric for visualisation. The plot with ground truth colouring reveals $10$ prominent clusters corresponding to each unique digit. However, several smaller groups are also observed indicating that the ground truth may not necessarily represent correct assignments since interpretation of the digits is sometimes ambiguous. The proposed clustering method is applied to the dataset after a UMAP transformation to 10 dimensions, and $5\%$ of the data ($90$ examples) is randomly sampled to serve as seeds for the clustering process. The clustering results are shown in Figure \ref{fig:digits_umap_nassir}, where it achieves top tier performance according to the external validation metrics. However, a closer examination of the visual results also unveils an interesting finding: the algorithm not only successfully captures much of the primary clusters but also assigns points that may be outliers with respect to the clusters to the anomalous group. These anomalous parts of clusters or individual small clusters often represent either uniquely written digits, or slight variations within the same class, such as the digit $1$ with or without a horizontal baseline\footnote{The interactive plots where points in the graph can be visualised are available in the GitHub project at https://github.com/M-Nassir/clustering}. 

\subsubsection{20 Newsgroups Corpus}
\begin{figure}
	\includegraphics[width=\linewidth]{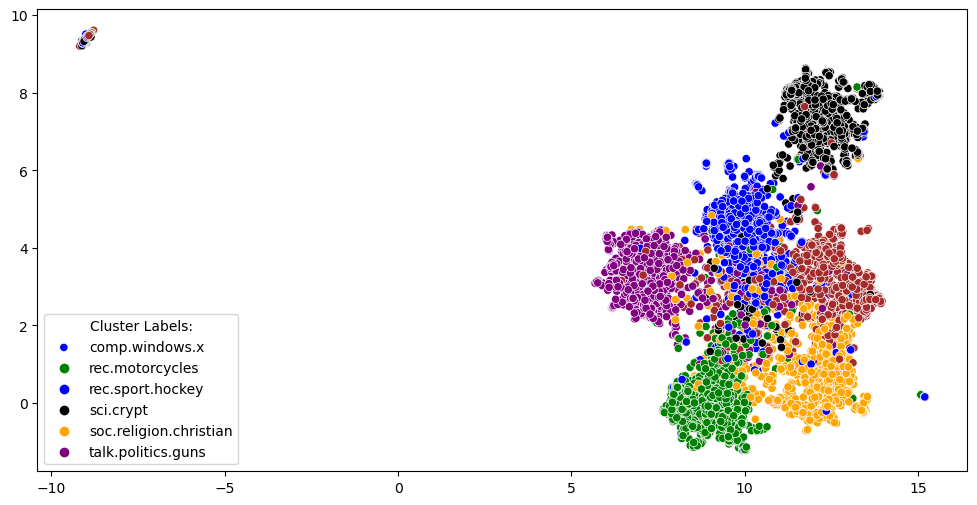}
	\caption{A 2-dimensional UMAP plot of the Newsgroup data set is shown. Each point is coloured according to its ground truth label.}
	\label{fig:20newsgrp_umap_labelled}
\end{figure}

The 20 Newsgroups dataset in the scikit-learn library is often used as an example of clustering data in the text document space. It contains over $18,000$ news articles that are organised into $20$ categories. In this specific example, a subset of $6$ categories is used. These were selected for their distinctness, ease of visualisation and interpretation, because many of the labels have been reported to be noisy, and hence warrant possibly belonging to other classes or could equally belong to multiple classes. 
The UMAP reduction to $2$ dimensions is shown in Figure \ref{fig:20newsgrp_umap_labelled}, where each example is coloured according to its ground truth category label. Notably, there is significant overlap between clusters, particularly in the central regions, and a small cluster consisting of examples from all categories can be observed on the far top-left of the plot.

\begin{figure}
	\includegraphics[width=\linewidth]{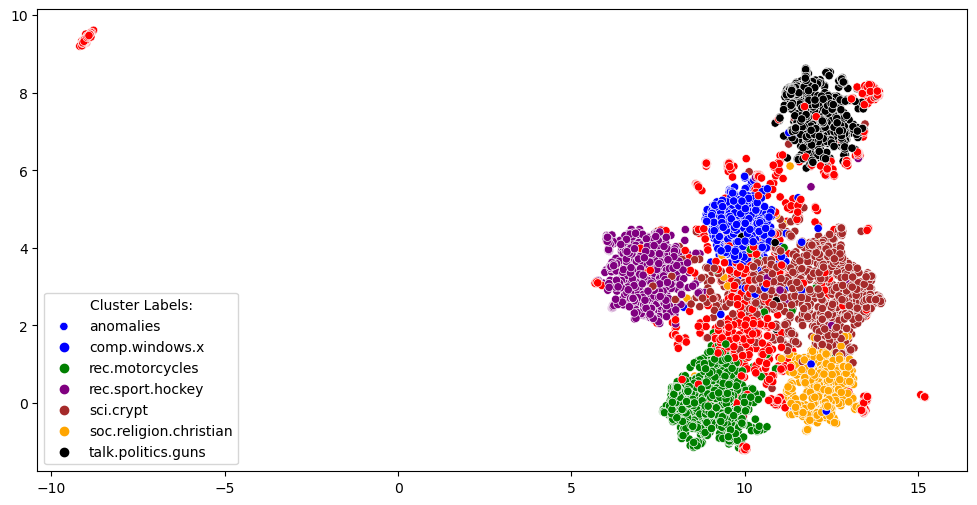}
	\caption{The results of applying the proposed semi-supervised clustering algorithm on $6$ of the $20$ Newsgroup corpus dataset categories after a UMAP reduction to $10$ dimensions.}
	\label{fig:20newsgrp_umap_nassir}
\end{figure}

For clustering purposes, the data is mapped to $10$ dimensions and a $1\%$ random sample of labelled examples is used as seeds to guide the algorithm. The clustering results are shown in Figure \ref{fig:20newsgrp_umap_nassir} (displayed over the 2-dimensional mapping). The algorithm successfully recovers most of the six main clusters, achieving top-tier scores according to the external validation metrics. Moreover, the algorithm flags many examples located between clusters as anomalies, as well as the small dense cluster on the far top-left. This again highlights that the algorithm performs a clustering as opposed to a partitioning of the data points, where anomalies are also detected.
Table \ref{table:cluster_keywords} provides the top $10$ keywords for each of the six clusters obtained by the proposed method, along with those from the anomalous group. The top keywords align well with the category labels, while the anomalous group consists of seemingly random words that do not align well to any specific category.

\begin{table}
	\centering
	\caption{Top Cluster Keywords}
	\label{table:cluster_keywords}
	\small 
	\begin{tabular}{|c|p{0.7\textwidth}|} 
	\hline
	\textbf{Category} & \textbf{Keywords} \\
	\hline
	\textbf{Anomalies} & helmet, thanks, edu, just, like, list, mail, com, dog, wa \\
	\hline
	\textbf{comp.windows.x} & use, display, application, program, motif, widget, thanks, file, server, window \\
	\hline
	\textbf{rec.motorcycles} & rider, road, just, like, riding, dod, motorcycle, ride, wa, bike \\
	\hline
	\textbf{rec.sport.hockey} & espn, season, playoff, play, year, player, hockey, wa, team, game \\
	\hline
	\textbf{sci.crypt} & escrow, use, nsa, algorithm, phone, government, encryption, clipper, chip, key \\
	\hline
	\textbf{soc.religion.christian} & christians, bible, people, sin, christ, christian, church, jesus, wa, god \\
	\hline
	\textbf{talk.politics.guns} & just, weapon, government, law, did, right, fbi, people, wa, gun \\
	\hline
	\end{tabular}
\end{table}	

\section{Conclusion}
\label{section:conclusion}
The present work introduces two significant contributions. Firstly, inspired by David Marr's tri-level hypothesis for information processing, it presents a computable definition of a cluster. This definition is derived by analysing elements of human visual clustering and assuming certain constraints deemed true of the real world. A key observation is that a cluster is a collection of points devoid of any anomalies with respect to a given grouping principle and measure. This sheds light on the dual nature of clustering and anomaly detection, highlighting their interdependence and the role of data representations and measures in achieving desired groupings.

Secondly, a novel semi-supervised clustering algorithm is proposed whose kernel is based on the anomaly detection work by Mohammad \cite{nassir2021anomaly}. This work gave a precise computable definition of what constitutes an anomaly and provides a compelling rationale for accepting such a definition. Additionally, an efficient algorithm to compute anomalies based on Expectation and Euclidean distance from the median is provided. The clustering algorithm operates on numerical data and utilises a small amount of labelled examples for initialisation, guidance, and subsequent expansion of clusters. Results have shown that as a rule of thumb, $10$-$30$ points per known cluster, give effective clustering results. The algorithm handles noisy labelled data by ejecting unexpected points (anomalies) from the groups. Instead of aiming for a complete partitioning of all the data, the algorithm focuses on clustering the data, thus ensuring that anomalies and unknown clusters are not forcibly assigned to existing clusters. If full partitioning is required, points can optionally be assigned to their nearest cluster based on anomaly scores. Such instances are left to be analysed in subsequent steps. Crucially, the proposed clustering algorithm is parameter-free, fast, and efficient—eliminating the need for users to manually set parameters.

The qualitative analysis over synthetic datasets and subsequent quantitative evaluation using established external validation metrics demonstrated the effectiveness of the proposed algorithm to consistently obtain top-tier or competitive performance. The results thus indicate its potential for real-world applications and as a novel contribution to the field of clustering. The use of dimensionality reduction techniques—particularly UMAP—in combination with clustering has also proven beneficial, as demonstrated in both image and text data analyses. The adopted workflow, which integrates dimensionality reduction for visualisation with sampling, labelling, and clustering, enhances cluster identification while also improving overall efficiency.

Clustering is an interactive and iterative process that combines exploration and analysis. Merely applying an algorithm without understanding the data and the problem space is not recommended, and results are never accepted blindly. During the exploratory phase, it is important to gain familiarity with the data and establish certain assumptions that constrain solutions, such as the type and number of clusters, the possibility of clusters forming hyper-spherical groupings, algorithm parameters, or the labelling of points for seeding and guiding an algorithm. In the presented semi-supervised clustering algorithm, the explicit constraints are taken to be labelled examples as opposed to parameter specification, and implicit is the choice of grouping law and measure. However, note the difference between knowledge of a few labelled examples per cluster and estimating parameters such as the $k$ in $k$-means or $mps$ and $eps$ in DBSCAN. In the former case clusters can be gradually built over the labelled subset, and then expanded upon, without requiring knowledge of samples from \emph{all} the clusters or requiring parameter estimation.

The presented work highlights how all algorithms constrain solutions, thereby reducing the number of possible groupings that can be returned. However, it is postulated that only relatively few grouping principles are of actual interest. These principles and appropriate measures constrain what groupings are acceptable a-prori, particularly when formulated through representation and unexpectedness. Hence, a realist position is adopted, where clustering is deemed an objective process as opposed to a purely subjective experience.

This work highlights the need to explore alternative grouping principles and similarity measures beyond proximity and Euclidean distance to the median—particularly for more effective anomaly detection within clusters using Expectations. Broadening these foundations could enhance the robustness and adaptability of the clustering approach. As the semi-supervised algorithm relies on a small set of labelled data to initialise and guide the clustering process, future work will also investigate methods for automatically generating reliable seed points, with the goal of achieving full automation. While the algorithm’s speed and design make it well-suited for online clustering, performance profiling has identified key bottlenecks, including excessive iterations during the clustering and anomaly ejection phases, as well as areas for optimisation in the underlying anomaly detection mechanism. Future work will address the performance isses of the algorithm and implementation.


\bibliography{Bibliography.bib} 

\begin{thebibliography}{33}
\providecommand{\natexlab}[1]{#1}
\providecommand{\url}[1]{\texttt{#1}}
\expandafter\ifx\csname urlstyle\endcsname\relax
  \providecommand{\doi}[1]{doi: #1}\else
  \providecommand{\doi}{doi: \begingroup \urlstyle{rm}\Url}\fi

\bibitem[Babaki(2017)]{behrouz_babaki_2017_831850}
Behrouz Babaki.
\newblock Cop-kmeans version 1.5, July 2017.
\newblock URL \url{https://doi.org/10.5281/zenodo.831850}.

\bibitem[Basu et~al.(2002)Basu, Banerjee, and Mooney]{BasuBM02}
Sugato Basu, Arindam Banerjee, and Raymond~J. Mooney.
\newblock Semi-supervised clustering by seeding.
\newblock In \emph{{ICML}}, pages 27--34. Morgan Kaufmann, 2002.

\bibitem[Basu et~al.(2004)Basu, Banerjee, and Mooney]{basu_pckmeans_2004}
Sugato Basu, Arindam Banerjee, and Raymond~J. Mooney.
\newblock Active semi-supervision for pairwise constrained clustering.
\newblock \emph{Proceedings of the tenth ACM SIGKDD international conference on
  Knowledge discovery and data mining}, pages 333--342, 2004.

\bibitem[Beer et~al.(2024)Beer, Weber, Miklautz, Leiber, Durani, B{\"{o}}hm,
  and Plant]{SHADE_2024}
Anna Beer, Pascal Weber, Lukas Miklautz, Collin Leiber, Walid Durani, Christian
  B{\"{o}}hm, and Claudia Plant.
\newblock {SHADE:} deep density-based clustering.
\newblock In \emph{{ICDM}}, pages 675--680. {IEEE}, 2024.

\bibitem[Bilenko et~al.(2004)Bilenko, Basu, and Mooney]{BilenkoBM04}
Mikhail Bilenko, Sugato Basu, and Raymond~J. Mooney.
\newblock Integrating constraints and metric learning in semi-supervised
  clustering.
\newblock In \emph{{ICML}}, volume~69 of \emph{{ACM} International Conference
  Proceeding Series}. {ACM}, 2004.

\bibitem[Bradley et~al.(2000)Bradley, Bennett, and
  Demiriz]{constrained_kmeans_bradley}
P.~Bradley, Kristin~P. Bennett, and Ayhan Demiriz.
\newblock Constrained k-means clustering.
\newblock Technical Report MSR-TR-2000-65, Microsoft Research, August 2000.

\bibitem[Chawla and Gionis(2013)]{ChawlaGionis2013}
Sanjay Chawla and Aristides Gionis.
\newblock k-means-: {A} unified approach to clustering and outlier detection.
\newblock In \emph{{SDM}}, pages 189--197. {SIAM}, 2013.

\bibitem[Desolneux et~al.(2007)Desolneux, Moisan, and Morel]{Morel07}
Agns Desolneux, Lionel Moisan, and Jean-Michel Morel.
\newblock \emph{From Gestalt Theory to Image Analysis: A Probabilistic
  Approach}.
\newblock Springer Publishing Company, Incorporated, 1st edition, 2007.
\newblock ISBN 0387726357.

\bibitem[Ester et~al.(1996)Ester, Kriegel, Sander, and Xu]{ester1996density}
Martin Ester, Hans-Peter Kriegel, Jörg Sander, and Xiaowei Xu.
\newblock A density-based algorithm for discovering clusters in large spatial
  databases with noise.
\newblock \emph{Proceedings of the 2nd International Conference on Knowledge
  Discovery and Data Mining (KDD)}, pages 226--231, 1996.

\bibitem[Forgy(1965)]{forgy65}
E.~Forgy.
\newblock Cluster analysis of multivariate data: Efficiency versus
  interpretability of classification.
\newblock \emph{Biometrics}, 21\penalty0 (3):\penalty0 768--769, 1965.

\bibitem[Gonz{\'{a}}lez{-}Almagro et~al.(2021)Gonz{\'{a}}lez{-}Almagro, Luengo,
  Cano, and Garc{\'{\i}}a]{GeneticGonzalezAlmagro21}
Germ{\'{a}}n Gonz{\'{a}}lez{-}Almagro, Juli{\'{a}}n Luengo,
  Jos{\'{e}}~Ram{\'{o}}n Cano, and Salvador Garc{\'{\i}}a.
\newblock Enhancing instance-level constrained clustering through differential
  evolution.
\newblock \emph{Appl. Soft Comput.}, 108:\penalty0 107435, 2021.

\bibitem[Gonz{\'{a}}lez{-}Almagro et~al.(2025)Gonz{\'{a}}lez{-}Almagro,
  Peralta, Poorter, Cano, and Garc{\'{\i}}a]{GonzalezAlmagroPPCG25}
Germ{\'{a}}n Gonz{\'{a}}lez{-}Almagro, Daniel Peralta, Eli~De Poorter,
  Jos{\'{e}}~Ram{\'{o}}n Cano, and Salvador Garc{\'{\i}}a.
\newblock Semi-supervised constrained clustering: an in-depth overview, ranked
  taxonomy and future research directions.
\newblock \emph{Artif. Intell. Rev.}, 58\penalty0 (5):\penalty0 157, 2025.

\bibitem[Kamvar et~al.(2003)Kamvar, Klein, and Manning]{KamvarKM03}
Sepandar~D. Kamvar, Dan Klein, and Christopher~D. Manning.
\newblock Spectral learning.
\newblock In \emph{{IJCAI}}, pages 561--566. Morgan Kaufmann, 2003.

\bibitem[Kelly et~al.(2025)Kelly, Longjohn, and Nottingham]{kelly2025uci}
Markelle Kelly, Rachel Longjohn, and Kolby Nottingham.
\newblock The {UCI} machine learning repository.
\newblock \url{https://archive.ics.uci.edu}, 2025.
\newblock Accessed July 2025.

\bibitem[Leiber et~al.(2023)Leiber, Miklautz, Plant, and
  Böhm]{leiber2023benchmarking}
Collin Leiber, Lukas Miklautz, Claudia Plant, and Christian Böhm.
\newblock Benchmarking deep clustering algorithms with clustpy.
\newblock In \emph{2023 IEEE International Conference on Data Mining Workshops
  (ICDMW)}, pages 625--632. IEEE, 2023.
\newblock \doi{10.1109/ICDMW60847.2023.00087}.

\bibitem[Lelis and Sander(2009)]{SSDBSCANLelisS09}
Levi Lelis and J{\"{o}}rg Sander.
\newblock Semi-supervised density-based clustering.
\newblock In \emph{{ICDM}}, pages 842--847. {IEEE} Computer Society, 2009.

\bibitem[Levy-Kramer(2018)]{Levy-Kramer_k-means-constrained_2018}
Josh Levy-Kramer.
\newblock {k-means-constrained}, April 2018.
\newblock URL \url{https://github.com/joshlk/k-means-constrained}.

\bibitem[Liu et~al.(2021)Liu, Li, Wu, and Fu]{LiuCOR2018}
Hongfu Liu, Jun Li, Yue Wu, and Yun Fu.
\newblock Clustering with outlier removal.
\newblock \emph{{IEEE} Trans. Knowl. Data Eng.}, 33\penalty0 (6):\penalty0
  2369--2379, 2021.

\bibitem[MacQueen(1967)]{macqueen1967}
J.~MacQueen.
\newblock Some methods for classification and analysis of multivariate
  observations.
\newblock In \emph{Proceedings of the Fifth Berkeley Symposium on Mathematical
  Statistics and Probability, Volume 1: Statistics}, pages 281--297, Berkeley,
  Calif., 1967. University of California Press.

\bibitem[Marr(1982)]{Mar82}
David Marr.
\newblock \emph{Vision: A Computational Investigation into the Human
  Representation and Processing of Visual Information}.
\newblock Henry Holt and Co., Inc., USA, 1982.
\newblock ISBN 0716715678.

\bibitem[McInnes et~al.(2018)McInnes, Healy, and Melville]{mcinnes2018umap}
Leland McInnes, John Healy, and James Melville.
\newblock Umap: Uniform manifold approximation and projection for dimension
  reduction.
\newblock \emph{arXiv preprint arXiv:1802.03426}, 2018.

\bibitem[Mohammad(2021)]{nassir2021anomaly}
Nassir Mohammad.
\newblock Anomaly detection using principles of human perception, 2021.

\bibitem[Pedregosa et~al.(2011)Pedregosa, Varoquaux, Gramfort, Michel, Thirion,
  Grisel, Blondel, Prettenhofer, Weiss, Dubourg, Vanderplas, Passos,
  Cournapeau, Brucher, Perrot, and Duchesnay]{scikitlearn}
F.~Pedregosa, G.~Varoquaux, A.~Gramfort, V.~Michel, B.~Thirion, O.~Grisel,
  M.~Blondel, P.~Prettenhofer, R.~Weiss, V.~Dubourg, J.~Vanderplas, A.~Passos,
  D.~Cournapeau, M.~Brucher, M.~Perrot, and E.~Duchesnay.
\newblock Scikit-learn: Machine learning in {P}ython.
\newblock \emph{Journal of Machine Learning Research}, 12:\penalty0 2825--2830,
  2011.

\bibitem[Peng et~al.(2016)Peng, Xiao, Feng, Yau, and Yi]{PengXFYY16}
Xi~Peng, Shijie Xiao, Jiashi Feng, Wei{-}Yun Yau, and Zhang Yi.
\newblock Deep subspace clustering with sparsity prior.
\newblock In \emph{{IJCAI}}, pages 1925--1931. {IJCAI/AAAI} Press, 2016.

\bibitem[Ren et~al.(2019)Ren, Hu, Dai, Pan, Hoi, and Xu]{RenHDPHX19}
Yazhou Ren, Kangrong Hu, Xinyi Dai, Lili Pan, Steven C.~H. Hoi, and Zenglin Xu.
\newblock Semi-supervised deep embedded clustering.
\newblock \emph{Neurocomputing}, 325:\penalty0 121--130, 2019.

\bibitem[Shental et~al.(2003)Shental, Bar{-}Hillel, Hertz, and
  Weinshall]{ShentalBHW03}
Noam Shental, Aharon Bar{-}Hillel, Tomer Hertz, and Daphna Weinshall.
\newblock Computing gaussian mixture models with {EM} using equivalence
  constraints.
\newblock In \emph{{NIPS}}, pages 465--472. {MIT} Press, 2003.

\bibitem[Tian et~al.(2014)Tian, Gao, Cui, Chen, and Liu]{TianGCCL14}
Fei Tian, Bin Gao, Qing Cui, Enhong Chen, and Tie{-}Yan Liu.
\newblock Learning deep representations for graph clustering.
\newblock In \emph{{AAAI}}, pages 1293--1299. {AAAI} Press, 2014.

\bibitem[Wagstaff et~al.(2001)Wagstaff, Cardie, Rogers, and
  Schrödl]{WagstaffCRS01}
Kiri Wagstaff, Claire Cardie, Seth Rogers, and Stefan Schrödl.
\newblock Constrained k-means clustering with background knowledge.
\newblock In \emph{ICML}, pages 577--584, 2001.

\bibitem[Ward~Jr(1963)]{Ward63}
Joe~H Ward~Jr.
\newblock Hierarchical grouping to optimize an objective function.
\newblock \emph{Journal of the American Statistical Association}, 58\penalty0
  (301):\penalty0 236--244, 1963.

\bibitem[Xie et~al.(2016)Xie, Girshick, and Farhadi]{XieGF16_DEC}
Junyuan Xie, Ross~B. Girshick, and Ali Farhadi.
\newblock Unsupervised deep embedding for clustering analysis.
\newblock In \emph{{ICML}}, volume~48 of \emph{{JMLR} Workshop and Conference
  Proceedings}, pages 478--487. JMLR.org, 2016.

\bibitem[Yang et~al.(2017)Yang, Fu, Sidiropoulos, and Hong]{Yang2017_DCN}
Bo~Yang, Xiao Fu, Nicholas~D. Sidiropoulos, and Mingyi Hong.
\newblock Towards k-means-friendly spaces: Simultaneous deep learning and
  clustering.
\newblock In \emph{{ICML}}, volume~70 of \emph{Proceedings of Machine Learning
  Research}, pages 3861--3870. {PMLR}, 2017.

\bibitem[Yu and Shi(2001)]{YuS01}
Stella~X. Yu and Jianbo Shi.
\newblock Grouping with bias.
\newblock In \emph{{NIPS}}, pages 1327--1334. {MIT} Press, 2001.

\bibitem[Zhang et~al.(2019)Zhang, Basu, and Davidson]{ZhangBD19}
Hongjing Zhang, Sugato Basu, and Ian Davidson.
\newblock A framework for deep constrained clustering - algorithms and
  advances.
\newblock In \emph{{ECML/PKDD} {(1)}}, volume 11906 of \emph{Lecture Notes in
  Computer Science}, pages 57--72. Springer, 2019.

\end{thebibliography}
\bibliographystyle{plainnat}

\end{document}